\documentclass{article}

\usepackage[final]{neurips_2021}
\usepackage{microtype}
\usepackage{makecell}
\usepackage{graphicx}
\usepackage{subfigure}
\usepackage{booktabs} 
\usepackage{stfloats}

\usepackage[utf8]{inputenc} 
\usepackage[T1]{fontenc}    
\usepackage{hyperref}       
\usepackage{url}            
\usepackage{booktabs}       
\usepackage{amsfonts}       
\usepackage{nicefrac}       
\usepackage{microtype}      
\usepackage{xcolor}         

\usepackage{mathrsfs}

\newcommand{\rom}[1]{\MakeUppercase{\romannumeral #1}}

\usepackage{graphicx}
\usepackage{floatflt} 
\usepackage{paralist} 
\usepackage{algorithm} 
\usepackage{amsthm}
\usepackage{amssymb}
\usepackage{listings}    
\usepackage{xcolor}
\usepackage{hyperref}
\usepackage{indentfirst}
\usepackage{booktabs}
\usepackage{subfigure}
\usepackage{caption}
\usepackage{calligra}
\usepackage{fancyhdr}
\usepackage{float}
\usepackage{pifont}
\usepackage{colortbl} 
\usepackage{amsmath}
\usepackage{MnSymbol}%
\usepackage{amsbsy}
\usepackage{multirow}

\def\x{\mathbf{x}}
\def\y{\mathbf{y}}
\def\z{\mathbf{z}}

\def\s{\mathbf{s}}
\def\h{\mathbf{h}}
\def\w{\boldsymbol{w}}

\def\LL{\mathcal{L}}
\def\FF{\mathcal{F}}

\def\D{\mathcal{D}}
\def\R{\mathbb{R}}
\def\NN{\mathbb{N}}
\def\N{\mathcal{N}}

\DeclareMathOperator{\X}{X}
\DeclareMathOperator{\Y}{Y}

\title{Collaborative Uncertainty  \\
in Multi-Agent Trajectory Forecasting}

\author{Bohan Tang$^1$ \quad Yiqi Zhong$^{2}$ \quad Ulrich Neumann$^2$ \quad Gang Wang$^{3}$ \quad Ya Zhang$^{1}$ \quad Siheng Chen$^{1}$\thanks{The corresponding author is Siheng Chen.}\\
$^{1}$Shanghai Jiao Tong University \quad $^{2}$University of Southern California \quad $^{3}$Beijing Institute of Technology \\ 
$^{1}$\texttt{tangbohan@alumni.sjtu.edu.cn}  \quad
$^{1}$\texttt{\{sihengc, ya\_zhang\}@sjtu.edu.cn} \\ \quad 
$^{2}$\texttt{\{yiqizhon, uneumann\}@usc.edu} $^{3}$\texttt{gangwang@bit.edu.cn} \\
}

\begin{document}

\maketitle

\begin{abstract}
Uncertainty modeling is critical in trajectory forecasting systems for both interpretation and safety reasons. To better predict the future trajectories of multiple agents, recent works have introduced interaction modules to capture interactions among agents. This approach leads to correlations among the predicted trajectories. However, the uncertainty brought by such correlations is neglected. To fill this gap, we propose a novel concept, \emph{collaborative uncertainty} (CU), which models the uncertainty resulting from the interaction module. We build a general CU-based framework to make a prediction model learn the future trajectory and the corresponding uncertainty. The CU-based framework is integrated as a plugin module to current state-of-the-art (SOTA) systems and deployed in two special cases based on multivariate Gaussian and Laplace distributions. In each case, we conduct extensive experiments on two synthetic datasets and two public, large-scale benchmarks of trajectory forecasting. The results are promising: 1) The results of synthetic datasets show that CU-based framework allows the model to appropriately approximate the ground-truth distribution. 2) The results of trajectory forecasting benchmarks demonstrate that the CU-based framework steadily helps SOTA systems improve their performances. Specially, the proposed CU-based framework helps VectorNet improve by $57$ cm regarding Final Displacement Error on nuScenes dataset. 3) The visualization results of CU illustrate that the value of CU is highly related to the amount of the interactive information among agents.
\end{abstract}

\vspace{-.8em}
\section{Introduction}
\label{introduction}
\vspace{-0.5em}
A multi-agent trajectory forecasting system aims to predict future trajectories of multiple agents based on their observed trajectories and surroundings~\cite{DBLP:journals/thms/StahlDJ14,2016Supporting}. Precise trajectory prediction provides essential information for decision making and safety in numerous intelligent systems, including autonomous vehicles~\cite{liang2020learning,Gao_2020_CVPR,Ye_2021_CVPR,gilles2021home}, drones~\cite{DBLP:conf/icc/XiaoZHY19}, and industrial robotics~\cite{DBLP:conf/icml/JetchevT09,DBLP:conf/aimech/RosmannO0B17}.

The rapid development of deep learning has enabled a number of deep-learning-based algorithms to handle multi-agent trajectory forecasting~\cite{liang2020learning,Gao_2020_CVPR,Ye_2021_CVPR,gilles2021home,Zhao_2019_CVPR,Choi_2019_ICCV,salzmann2020trajectron++,zeng2021lanercnn,li2020evolvegraph,DBLP:conf/nips/KosarajuSM0RS19}. These methods exhibit state-of-the-art performances, with some having been integrated into real-world systems. However, deep-learning-based forecasting is not always reliable or interpretable~\cite{Gal2016Uncertainty,NIPS2017_2650d608,zhao2020uncertainty}. In circumstances when noises from the environment are overwhelmingly distracting, or when the situation has never been encountered before, a deep-learning-based algorithm could provide baffling predictions, which might cause terrible tragedies. A fundamental challenge is to know when we could rely on those deep-learning-based forecasting algorithms. To tackle this problem, one solution is to report the uncertainty of each prediction. Finding ways to best conceptualize and measure the prediction uncertainty of deep-learning-based algorithms becomes an imperative, which motivates this work.

\begin{figure}[t]
\vspace{-1em}
\begin{center}
\includegraphics[width=1.\textwidth]{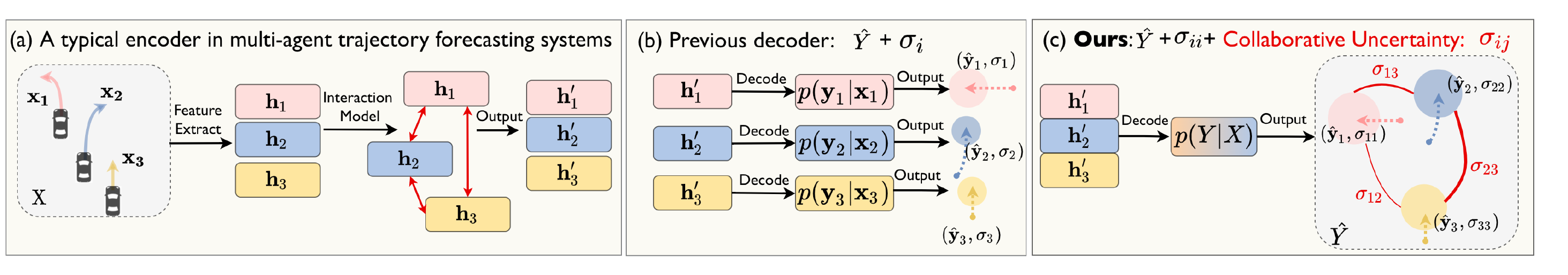}
\vspace{-2em}
\caption{\textbf{Uncertainty modeling in multi-agent trajectory forecasting.} (a) a typical pipeline of an encoder in multi-agent trajectory forecasting systems. (b) and (c) illustrate the decoder pipeline of previous methods and our method respectively. Previous methods output the predicted trajectory $\hat{Y}$ and individual uncertainty $\sigma_i$, and our method additionally outputs \textit{collaborative uncertainty} $\sigma_{ij}$.}
\label{example_out}
\vspace{-8mm}
\end{center}
\end{figure}

There are two main types of uncertainty to model in deep-learning-based algorithms~\cite{der2009aleatory}: 1) aleatoric uncertainty, regarding information aside from statistic models, which data cannot explain; 2) epistemic uncertainty, the uncertainty inside a model, when the model lacks knowledge of the system/process being modeled (e.g., due to limited training data). As it is most effective to model aleatoric uncertainty in big data regimes such as those common to deep learning with image data~\cite{NIPS2017_2650d608}, this work focuses on aleatoric uncertainty. In the following passage, we use the term ``uncertainty'' to represent aleatoric uncertainty.  \cite{Gal2016Uncertainty} uses the predictive variance to approximate uncertainty in the Bayesian deep learning model, which has been widely adapted in many works~\cite{DBLP:conf/cvpr/BhattacharyyaFS18,DBLP:conf/corl/Jain0LXFSU19,DBLP:conf/cvpr/HongSP19,Choi_2019_ICCV} for uncertainty modeling in multi-agent trajectory forecasting. However, the predictive variance of a single agent alone may not suffice to reflect the complete landscape of uncertainty, especially when agent-wise interaction is present. Recent works that attempt to exploit the interaction among agents have impressively boosted the prediction precision, which further highlights the need to better measure uncertainty in multi-agent trajectory forecasting. We seek to build a more sophisticated and robust measurement for capturing the previously neglected uncertainty brought by correlated predictions.

In this paper, we coin a concept \emph{individual uncertainty} (IU) to describe the uncertainty that can be approximated by the predictive variance of a single agent. Relatively, we propose a new concept, \emph{collaborative uncertainty} (CU) to estimate the uncertainty resulting from the usage of interaction modules in prediction models. We further introduce an original probabilistic CU-based framework to measure both individual and collaborative uncertainty in the multi-agent trajectory forecasting task. We apply this framework to two special cases: multivariate Gaussian distribution and multivariate Laplace distribution. In each case, our CU-based framework allows our model to simultaneously learn the mappings that are from input data to 1) accurate prediction, 2) individual uncertainty, and 3) collaborative uncertainty; see Figure~\ref{example_out} for model illustration. Extensive experiments demonstrate that CU modeling yields significantly larger performance gains in prediction models equipped with interaction modules (See Figure \ref{compare_a2a}), confirming that CU is highly related to the existence of the interaction modeling procedure, and adding CU modeling benefits accurate predictions.

The contributions of this work are summarized as follows:

$\bullet$ We propose, analyze, and visualize a novel concept, \textit{collaborative uncertainty} (CU), to model the uncertainty brought by the interaction modules in multi-agent trajectory forecasting.

$\bullet$  We design a general CU-based framework to empower a prediction model to generate a probabilistic output, where the mean is the future trajectory and the covariance reflects the corresponding uncertainty. Under this framework, we show two special cases based on multivariate Gaussian and Laplace distributions respectively.

$\bullet$ We conduct extensive experiments to validate the CU-empowered prediction model on both synthetic datasets and two large-scale real-world datasets. On self-generated synthetic datasets, we validate the proposed method is able to closely reconstruct the ground-truth distribution.  On the public benchmarks, the CU-empowered prediction model consistently outperforms the corresponding one without CU. Specially, by leveraging the proposed CU, VectorNet improves by $57$ cm regarding Final Displacement Error (FDE) on nuScenes dataset!

\vspace{-1em}
\section{Related Works}

\vspace{-1em}
\label{rw}
\textbf{Aleatoric uncertainty modeling in deep learning.} 
Recent efforts are rising as to improve the measurement of aleatoric uncertainty in deep learning models. One seminal work is \cite{NIPS2017_2650d608}. It proposes a unified Bayesian deep learning framework to explicitly represent aleatoric uncertainty using predictive variance for generic regression and classification tasks. Many existing works~\cite{DBLP:conf/cvpr/KendallGC18,ayhan2018test,pmlr-v80-depeweg18a,feng2018towards,wang2019aleatoric,pmlr-v119-kong20b} follow this idea and formulate uncertainty as learned loss attenuation. For example, to make predictive-variance-based aleatoric uncertainty measurements more efficient, \cite{ayhan2018test} adds data augmentation during the test-time. But, these works only pay attention to individual uncertainty. 
 
Other recent works attend to the uncertainty measurement for correlated predictive distributions. For example,~\cite{8578672} and \cite{DBLP:conf/nips/MonteiroFCPMKWG20} measure spatially correlated uncertainty in a generative model for respectively image reconstruction and pixel-wise classification, and~\cite{hafner2021mastering} captures joint uncertainty as discrete variables in the field of reinforcement learning. Despite these three works, our work is the first to conceptualize and measure collaborative uncertainty in the multi-agent trajectory forecasting task. To the best of our knowledge, there are only two papers~\cite{DBLP:conf/cvpr/GundavarapuSMSJ19,girgis2021autobots} close to our track. \cite{DBLP:conf/cvpr/GundavarapuSMSJ19} and \cite{girgis2021autobots} model the joint uncertainty in the pose estimation task and multi-agent trajectory forecasting task respectively. However, they present several limitations: 1) They only examined the circumstance where the model's output follows Gaussian distribution; 2) They did not provide a theoretical conceptualization or definition for the uncertainty due to correlated predictive distributions, and they did not analyze the causes of such uncertainty. These are essential problems to tackle. In this work, we not only formulate a general framework that works for both Gaussian and Laplace distributions, but we also theoretically conceptualize \emph{collaborative uncertainty} and analyze its causes.

\textbf{Multi-agent trajectory forecasting.} This task takes the observed trajectories from multiple agents as the inputs, and outputs the predicted trajectory for each agent. Like many other sequence prediction tasks, this task used to use a recurrent architecture to process the inputs~\cite{DBLP:conf/cvpr/AlahiGRRLS16,DBLP:conf/cvpr/HasanSTBGC18,DBLP:conf/cvpr/Liang0NH019}. Later, however, the graph neural networks become a more common approach as they can significantly assist trajectory forecasting by capturing the interactions among agents~\cite{liang2020learning,Gao_2020_CVPR,Ye_2021_CVPR,gilles2021home,salzmann2020trajectron++,zeng2021lanercnn,li2020evolvegraph,DBLP:conf/nips/KosarajuSM0RS19}. For safety reasons, it is necessary to report the uncertainty of each predicted trajectory. Works to date about uncertainty measurement ~\cite{Choi_2019_ICCV,DBLP:conf/corl/Jain0LXFSU19,DBLP:conf/cvpr/HongSP19,DBLP:conf/cvpr/LeeCVCTC17,DBLP:conf/eccv/FelsenLG18} have appropriately modeled the interaction among multi-agent trajectories for boosting performances, but they overlook the uncertainty resulting from the correlation in predicted trajectories. We seek to fill this gap by introducing and modeling \emph{collaborative uncertainty}.

\vspace{-1.5mm}
\section{Methodology}
\label{method}
\vspace{-1.5mm}
\subsection{Problem Formulation}
\vspace{-1.5mm}

Consider $m$ agents in a data sample, and let $\X\!=\!\{\!\x_{1}\!,\!\x_{2}\!,...,\!\x_{m}\!\}$, $\Y\!=\!\{\!\y_{1}\!,\!\y_{2}\!,\!...\!,\!\y_{m}\!\}$ be the past observed and the future trajectories of all agents, where $\x_{i}\!\in\!\R^{2T_{-}}$ and $\y_{i}\!\in\!\R^{2T_{+}}$ are the past observed and the future trajectories of the $i$-th agent. Each $\x_i/\y_i$ consists of two-dimensional coordinates at different timestamps of $T_{\!-}/T_{\!+}$. We assume that a training dataset $\D$ consists of $N$ individual and identically distributed data samples $\{(\!\X^{i}\!,\Y^{i}\!)\}_{i=1}^{N}$. For predicting future trajectories of multiple agents and modeling the uncertainty over the predictions, we seek to use a probabilistic framework to model the predictive distribution $p(\!\Y|\X\!)$ of multiple agents' future trajectories based on the training dataset $\D$.
\begin{figure}[h]
\vspace{-1.2em}
\begin{center}
\centerline{\subfigure[Individual Model]{
		\includegraphics[scale=0.3]{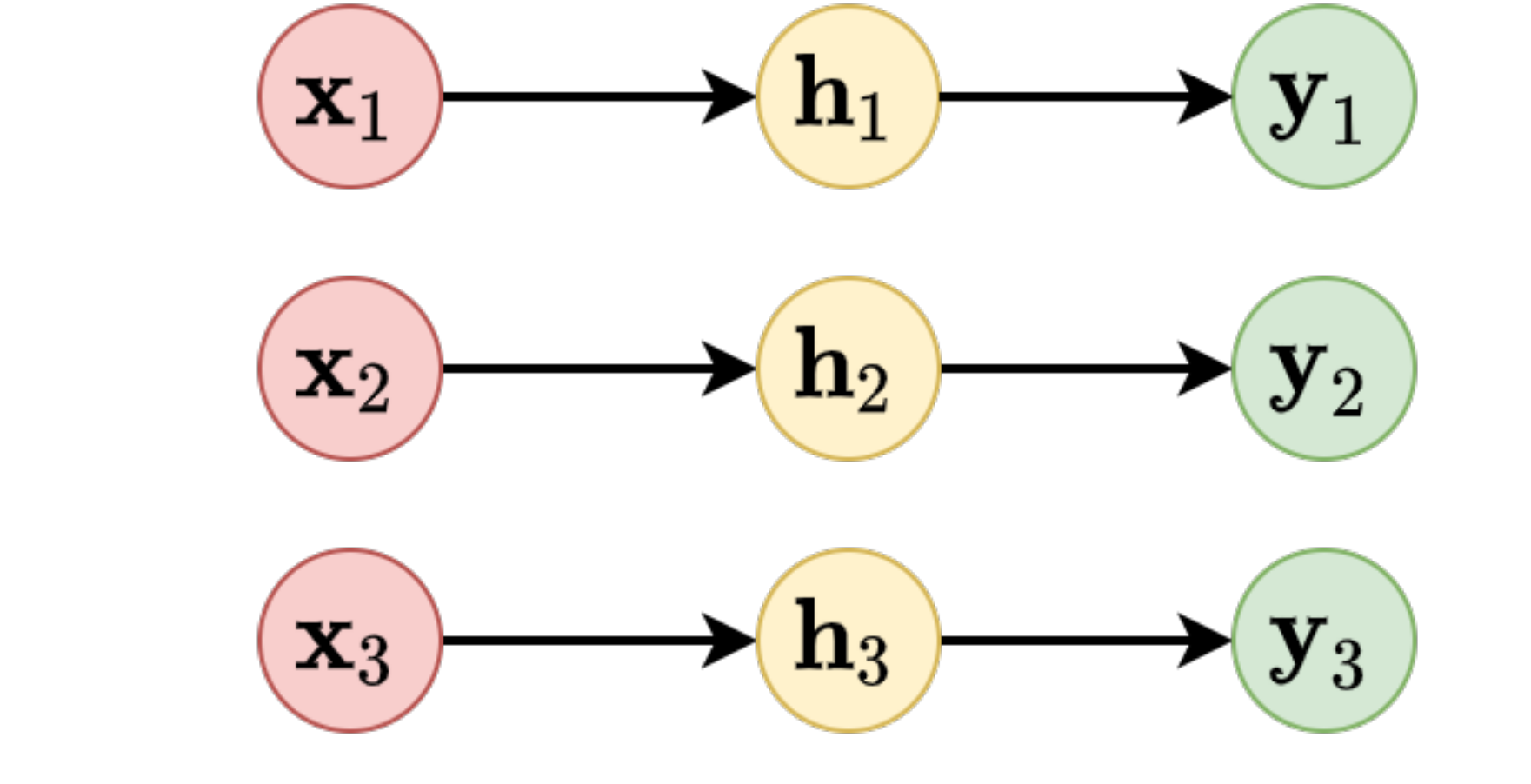}
        \label{fig:Individual}
	}
\hspace{2cm}
\subfigure[Collaborative Model]{
		\includegraphics[scale=0.15]{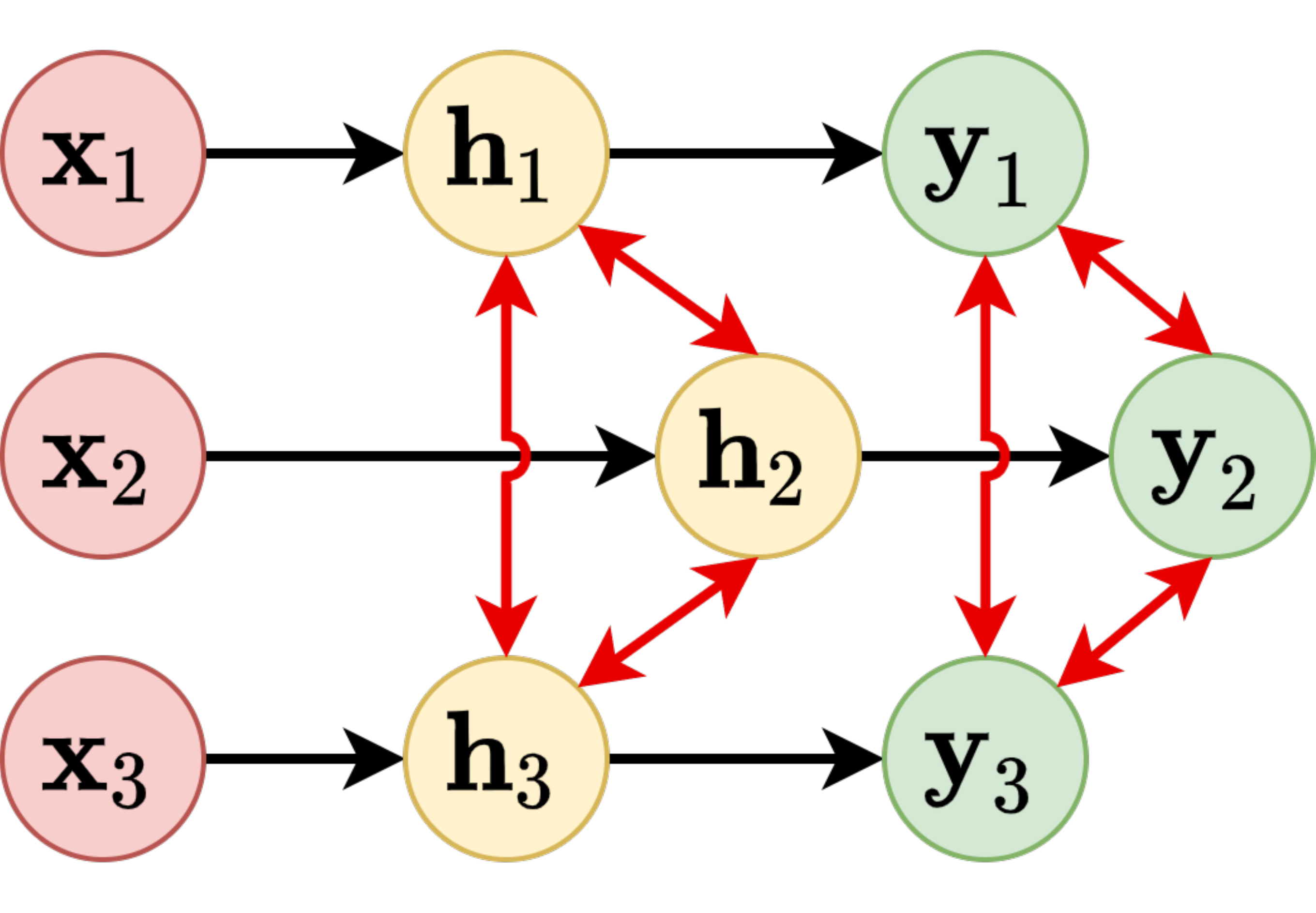}
        \label{fig:Collaborative}
	}
	}
\vspace{-3mm}
\caption{\textbf{Graphical model for deep learning networks in the three-agent trajectory forecasting setting}: (a) represents the model that predicts the trajectory of each agent independently; (b) shows the model that explicitly captures the interaction among multiple agents. $\x_{i}$ is the observed trajectory of the $i$-th agent; $\h_{i}$ and $\y_{i}$ are its corresponding hidden feature and future trajectory respectively.}
\vspace{-6mm}
\end{center}
\end{figure}
Previous works in uncertainty modeling~\cite{Gal2016Uncertainty,NIPS2017_2650d608,pmlr-v48-gal16} use Gaussian distribution to approximate $p(\!\Y|\X\!)$. The assumption behind this approach is that $p(\!\y_{i}|\x_{i}\!)$ is independent for every $i\in\{1,2,3,...,m\}$. Mathematically, they set the covariance matrix of $p(\!\Y|\X\!)$ as a diagonal matrix. This assumption is valid for the regression task that uses the model shown in Figure~\ref{fig:Individual}. We refer the uncertainty under the independence assumption as \emph{individual uncertainty} in this paper. However, Figure~\ref{fig:Collaborative} considers a prediction model that includes interaction modeling among multiple agents: $\y_{i}$ is no longer dependent solely on $\x_{i}$, but also on other agents $\x_{j}$ where $j\!\neq\!i$ in the scene. We call the uncertainty brought by this interaction \textit{collaborative uncertainty}. The existence of collaborative uncertainty turns $p(\!\Y|\X\!)$ from the individual distribution into the joint distribution of multiple agents. 

Contrary to existing methods, we consider collaborative uncertainty and model $p(\Y|\X)$ more accurately by making the covariance matrix a full matrix without imposing any restrictions on its form. In the following subsection, we will introduce an approach to modeling both individual uncertainty and collaborative uncertainty using a unified CU-based framework.

\begin{figure*}[ht]
\begin{center}
\centerline{\includegraphics[scale=0.45]{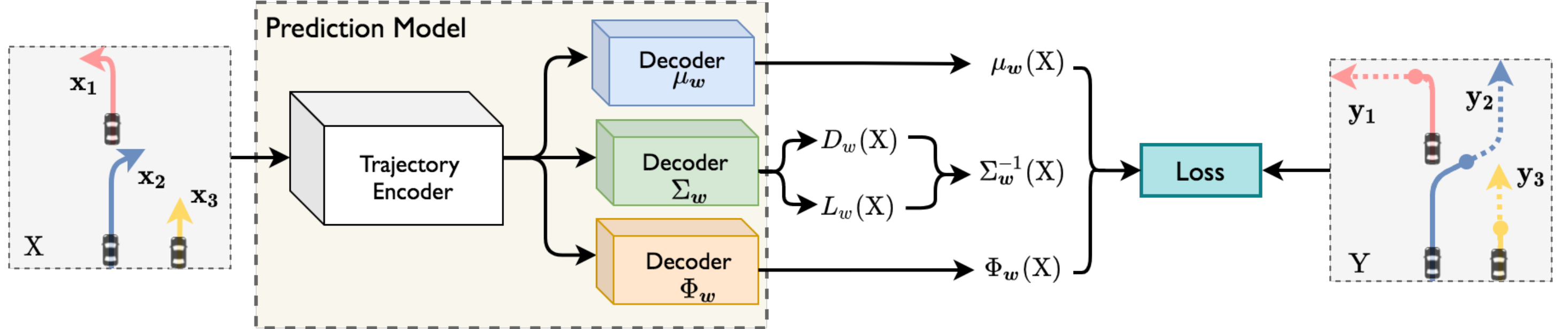}}
\vspace{-0.5em}
\caption{\textbf{Proposed uncertainty estimation framework.} The encoder may contain a module that exploits agent-wise interaction. Decoders output the mean $\mu_{\w}(\X)$, covariance $\Sigma_{\w}(\X)$ containing individual and collaborative uncertainty, and auxiliary parameters $\Phi_{\w}(\X)$. The outputs formulate the training loss with the ground truth $\Y$. $\Phi_{\w}(\cdot)$ is only used in Laplace collaborative uncertainty.}
\label{framework}
\end{center}
\vskip -0.38in
\end{figure*}

\subsection{General Formulation of Collaborative Uncertainty}
\label{3.2}
\vspace{-2mm}

In this work, to model collaborative uncertainty, we abandon the independence assumption held by previous works~\cite{Gal2016Uncertainty,NIPS2017_2650d608,pmlr-v48-gal16}, setting $p(\Y|\X)$ as a joint multivariate distribution, whose mean is $\mu \in \R^{m \times 2T_{+}}$ and covariance $\Sigma\in \R^{m\times m\times2T_{+}}$. Element $\mu_{i,t}$ is the expected position of the $i$-th agent at timestamp $t$. As the diagonal elements of $\Sigma$ are considered individual uncertainty~\cite{Gal2016Uncertainty,NIPS2017_2650d608,pmlr-v48-gal16}, we further let \textit{off-diagonal} elements describe collaborative uncertainty. Diagonal element $\Sigma_{i,i,t}$ models the variance of the $i$-th agent at timestamp $t$; off-diagonal element $\Sigma_{i,j,t}$ models the covariance between the $i$-th and $j$-th agents at timestamp $t$. Therefore, we can simultaneously obtain individual and collaborative uncertainty by estimating the covariance $\Sigma$ of $p(\Y|\X)$. Accordingly, we propose a CU-based comprehensive uncertainty estimation framework (see Figure~\ref{framework}) with the following steps:
\vspace{-0.5mm}

\textbf{Step 1:} \emph{Choose a probability density function,} $p(\!\Y|\X; \mu, \Sigma, \Phi\!)$, for the predictive distribution, which includes a mean $\mu\!\in\!\R^{\!m\!\times\!2T_{+}}$ used to approximate the future trajectories, a covariance $\Sigma\!\in\!\R^{\!m\times\!m\times\!2T_{+}}$ used to quantify individual uncertainty and collaborative uncertainty, and some auxiliary parameters $\Phi$ used to describe the predictive distribution. Further, we set covariance matrix $\Sigma_{t}$, which represents the covariance matrix at timestamp $t$, as a full matrix instead of an identity or diagonal matrix.
\vspace{-1.5mm}

\textbf{Step 2:} \emph{Design a prediction model,} $\FF[\mu_{\w}(\X),\Sigma_{\w}(\X),\Phi_{\w}(\X)]$, where $\mu_{\w}(\X)$, $\Sigma_{\w}(\X)$ and $\Phi_{\w}(\X)$ are three neural networks, which approximate values of mean $\mu$, covariance $\Sigma$ and auxiliary parameters $\Phi$ respectively. Note that $\w$ only indicates the parameters of these neural networks are trainable, and does not mean they share same parameters.
\vspace{-1.5mm}

\textbf{Step 3:} \emph{Derive a loss function} from $p(\Y|\X;\mu,\Sigma,\Phi)$ via maximum likelihood estimation:
$\LL(\w)=-\sum\limits_{i=1}^{N}\log p(\Y^{i}|\X^{i};\mu_{\w}(\X^{i}),\Sigma_{\w}(\X^{i}),\Phi_{\w}(\X^{i}))$ minimized to update trainable parameters in $\mu_{\w}(\cdot)$, $\Sigma_{\w}(\cdot)$ and $\Phi_{\w}(\cdot)$.

\vspace{-3mm}
\subsection{Two Special Cases}
\label{3.3}
\vspace{-4mm}
\begin{table*}[h]

\caption{Two special cases with various assumptions about covariance $\Sigma$. \textbf{DIA}: the diagonal matrix (individual uncertainty only). \textbf{FULL}: the full matrix (both individual and collaborative uncertainty).}
\label{com-table}
\vspace{-1em}
\begin{center}
\begin{footnotesize}
\begin{sc}
\begin{tabular}{c|c|c}
\toprule
\multirow{2}{*}{Assumption} &  \multicolumn{2}{c}{Loss Function of Two Special Cases}
\\
\cline{2-3}
&Gaussian Distribution & Laplace Distribution \\
\hline
DIA$:
\begin{pmatrix}
\sigma_{11} \!& \!0\!&\!\cdots\!&\!0\\
0\!&\!\sigma_{22}\!&\!\cdots\!&\!0\\
\vdots\!&\!\vdots\!&\!&\!\vdots  \\
0\!&\!0\!&\!\cdots\!&\!\sigma_{mm}  \\
\end{pmatrix}$\!

&{\scriptsize\!$\!\frac{1}{2}\! \sum\limits_{i=1}^{m}\![\sigma_{ii}^{-2}||\!\y_i\!-\!\mu_{\w}(\!\x_i\!)\!||_2^{2}\!+\!\log\!\sigma_{ii}^{2}]$\!}

&{\scriptsize\!$\!\sum\limits_{i=1}^{m}[\sigma_{ii}^{-2}||\!\y_i\!-\!\mu_{\w}(\!\x_i\!)\!||_1\!+\!\log\!\sigma_{ii}^{2}]$}
\\
\hline
\makecell[c]{FULL$:
\begin{pmatrix}
\sigma_{11}\!&\!\sigma_{12}\!&\!\cdots\!& \sigma_{1m}\\
\sigma_{21}\!&\!\sigma_{22}\!&\!\cdots\!&\!\sigma_{2m}  \\
\vdots\!&\!\vdots\!&\!&\!\vdots\\
\sigma_{m1}\!&\!\sigma_{m2}\!&\!\cdots\!&\!\sigma_{mm}\\
\end{pmatrix}$}\!

&{\scriptsize\!\makecell[c]{$\!\frac{1}{2} [q_{\w}(\Y,\X)\!-\sum\limits_{j=1}^{m}\!\log (d_{jj})]$} }

& {\scriptsize \makecell[c]{$ \frac{1}{2} [\frac{q_{\w}(\Y,\X)}{\Phi_{\w}(\X)}+m\log \Phi_{\w}(\X)-
    \!\sum\limits_{j=1}^{m}\!\log (d_{jj})]
    $}}
\\
\bottomrule
\end{tabular}
\end{sc}
\end{footnotesize}
\end{center}
\vspace{-0.15in}
\end{table*}
In multi-agent trajectory forecasting, based on Laplace and Gaussian distributions, the $\ell_1$- and $\ell_2$-based loss functions are commonly adopted to train prediction models~\cite{liang2020learning,Gao_2020_CVPR,DBLP:conf/nips/KosarajuSM0RS19,mangalam2020journey}. Here we apply the probabilistic framework proposed in Section~\ref{3.2} to model the individual and collaborative uncertainty based on multivariate Gaussian distribution and multivariate Laplace distribution respectively, which leads to two novel loss functions. Mathematically, the essential difference between our proposed loss functions and previous loss functions derived from Gaussian distribution and Laplace distribution for modeling individual uncertainty is that they have different assumptions about the covariance matrix; see a summary in Table~\ref{com-table}. We regard the covariance as a full matrix.

\vspace{-1.5mm}
\subsubsection{Gaussian collaborative uncertainty}
\label{g_cu}
\vspace{-1.5mm}

We start by the multivariate Gaussian distribution, as it has a simpler probability density function than the multivariate Laplace distribution.

\textbf{Probability density function. } We follow the framework proposed in Section~\ref{3.2} and choose the probability density function as the multivariate Gaussian distribution:
\begin{equation}
\setlength{\abovedisplayskip}{4pt}
\setlength{\belowdisplayskip}{4pt}
\begin{small}
p(\Y|\X;\mu,\Sigma,\Phi)=(2\pi)^{-\frac{m}{2}}\cdot {\rm det} [\Sigma]^{-\frac{1}{2}}\cdot
e^{ -\frac{1}{2}(\Y - \mu)\Sigma^{-1}(\Y - \mu)^{T}},
\label{g_pdf}
\end{small}
\end{equation}
where ${\rm det} [\Sigma]$ represents the determinant of covariance $\Sigma$. 

\textbf{Model design.} Based on (\ref{g_pdf}), we can approximate the value of mean $\mu$ via a neural network $\mu_{\w}(\cdot)$. When using the same way to approximate the value of covariance $\Sigma$, however, we face two challenges: 1) each covariance matrix $\Sigma_t$ in covariance $\Sigma$ needs to be inverted, which could lead to numerical instability; 2) it is computationally expensive and numerically unstable to compute the determinant of each covariance matrix $\Sigma_t$ in covariance $\Sigma$ directly given a large amount of trainable parameters. 

For the first challenge, we use a neural network $\Sigma^{-1}_{\!\w}(\cdot)$ to directly approximate the inverse of covariance $\Sigma$. For the second challenge, similar to \cite{8578672} and \cite{DBLP:conf/cvpr/GundavarapuSMSJ19}, we apply the square-root-free Cholesky decomposition to each $\Sigma^{-1}_{t_{\w}}$ in $\Sigma^{-1}_{\!\w}(\!\X\!)$: $\Sigma^{-1}_{\!\w}(\!\X\!)\!=\!L_{\!\w}(\!\X\!)D_{\!\w}(\!\X\!) L^{T}_{\!\w}(\!\X\!)$,

where $L_{\!\w}(\!\X\!)$ is a lower unit triangular matrix and $D_{\!\w}(\!\X\!)$ is a diagonal matrix. Then, the determinant of the inverse of covariance $\Sigma^{-1}$ is obtained by $\prod\limits_{j=1}^m\!d_{jj}$, where $d_{jj}$ is the $j$-th diagonal element in $D_{\!\w}(\!\X\!)$. We can thus get the parameterized form of ($\ref{g_pdf}$) as: $p(\!\Y\!|\!\X\!;\!\w\!)\!=\!(\!2\pi\!)^{-\!\frac{m}{2}}(\!\prod\limits_{j=1}^m\!d_{jj})^{\frac{1}{2}}\!e^{-\!\frac{q_{\w}(\!\Y\!,\!\X\!)}{2}},$
where $q_{\w}(\!\Y,\!\X\!)\!=\![\!\Y\!-\!\mu_{\w}(\!\X\!)\!]\Sigma^{-1}_{\w}(\!\X\!)[\!\Y\!-\!\mu_{\w}(\!\X\!)\!]^{T}$.

As there are no auxiliary parameters in the parameterized form of ($\ref{g_pdf}$), we can get the prediction model $\FF[\mu_{\w}(\X),\Sigma^{-1}_{\w}(\X)]$, whose framework is illustrated in Figure~\ref{framework}. Once $\Sigma^{-1}_{\w}(\X)$ is fixed and given, individual and collaborative uncertainty are computed through the inversion.

\textbf{Loss function. } According to the square-root-free Cholesky decomposition and the parameterized form of ($\ref{g_pdf}$), the Gaussian collaborative uncertainty loss function is then:
\begin{equation}
\setlength{\abovedisplayskip}{4pt}
\setlength{\belowdisplayskip}{4pt}
\begin{small}
\LL_{\rm Gau-cu}(\w)=\frac{1}{2} \frac{1}{N} \sum_{i=1}^{N}[q_{\w}(\Y^i,\X^i)-\sum_{j=1}^{m} \log (d^{i}_{jj})].
\label{loss_gau}
\end{small}
\end{equation}
We update the trainable parameters in $\mu_{\w}(\cdot)$ and $\Sigma^{-1}_{\w}(\cdot)$ through minimizing (\ref{loss_gau}). Note that $q_{\w}(\cdot,\cdot)$ is related to $\mu_{\w}(\cdot)$ and $\Sigma^{-1}_{\w}(\cdot)$, and $d^{i}_{jj}$ is related to $\Sigma^{-1}_{\w}(\cdot)$.

\vspace{-1.5mm}
\subsubsection{Laplace collaborative uncertainty}
\vspace{-1.5mm}

In multi-agent trajectory forecasting, previous methods~\cite{liang2020learning,Gao_2020_CVPR,DBLP:conf/nips/KosarajuSM0RS19} have found that the $\ell_1$-based loss function derived from Laplace distribution usually leads to better prediction performances than the $\ell_2$-based loss function from Gaussian distribution, because the former is more robust to outliers. It is thus important to consider multivariate Laplace distribution.

\textbf{Probability density function.}
We follow the framework proposed in Section~\ref{3.2} and choose the probability density function as the multivariate Laplace distribution:
\begin{equation}
\setlength{\abovedisplayskip}{4pt}
\setlength{\belowdisplayskip}{4pt}
\begin{small}
p(\Y|\X;\mu,\Sigma,\Phi) = \frac{2{\rm det} [\Sigma]^{-\frac{1}{2}} }{(2\pi)^{\frac{m}{2}}\lambda}\cdot\frac{K_{(\frac{m}{2}-1)}(\sqrt{\frac{2}{\lambda}(\Y\!-\!\mu)\Sigma^{-1}(\Y\!-\!\mu)^{T}})}{(\sqrt{\frac{\lambda}{2}(\Y\!-\!\mu)\Sigma^{-1}(\Y\!-\!\mu)^{T}})^{\frac{m}{2}}}, 
\label{la_pdf}
\end{small}
\end{equation}
where ${\rm det} [\!\Sigma\!]$ denotes the determinant of covariance $\Sigma$, and $K_{\!(\frac{m}{2}-1)\!}(\cdot)$ denotes the modified Bessel function of the second kind with order $(\!\frac{m}{2}-1\!)$.

\textbf{Model design.} Similar to Section~\ref{g_cu}, we employ two neural networks $\mu_{\w}(\cdot)$ and $\Sigma^{-1}_{\w}(\cdot)$ to approximate the values of $\mu$ and $\Sigma^{-1}$ respectively, and represent $\Sigma^{-1}_{\w}(\X)$ via its square-root-free Cholesky decomposition we used in the Gaussian collaborative uncertainty. Since the modified Bessel function is intractable for a neural network to work with, different from Section~\ref{g_cu}, we should simplify (\ref{la_pdf}).

Inspired by~\cite{1618702}, we simplify (\ref{la_pdf}) by utilizing the multivariate Gaussian distribution to approximate the multivariate Laplace Distribution. We reformulate a multivariate Laplace distribution by introducing  auxiliary variables. Let $\z\in\R^{+}$ be a random variable with the probability density function: $p(\z|\X;\w) = \frac{1}{\lambda}e^{-\frac{\z}{\lambda}}$, then we can get: $p(\!\Y|\z\!,\X;\w\!)\!=\!\frac{{\rm det}\![\!\Sigma^{-1}_{\w}\!(\X\!)\!]^{\frac{1}{2}}}{(2\pi\z)^{\frac{m}{2}}}e^{-\!\frac{q_{\w}(\Y,\X)}{2\z}}$,
where $q_{\w}(\Y,\X)=[\Y-\mu_{\w}(\X)]\Sigma^{-1}_{\w}(\X)[\Y-\mu_{\w}(\X)]^{T}$. Further, if the value of $\z$ is given, $p(\Y|\z,\X;\w)$ is a multivariate Gaussian distribution. In this work, instead of drawing a value for $\z$ from the exponential distribution, we use a neural network $\Phi_{\w}(\cdot)$ to directly output a value for $\z$. The intuition is that, in the training process of the prediction model, the value of $p(\Y|\X;\w)$ is the conditional expectation of $\z$ given $\X$ and $\Y$, which makes $p(\Y|\z,\X;w)$ a function of $\z$ whose domain is $\R^{+}$. Thus, there should exist an appropriate $\z^{*}\in\R^{+}$ to make: $p(\Y|\X;\w)=p(\Y|\z^{*},\X;\w)$ (see proof in the appendix). To find such a $\z^{*}$, we use $\Phi_{\w}(\X)$, which can employ its learning ability. Then, we can get the parameterized form of $p(\Y|\X;\w)$ as: $p(\Y|\X;\w) = \frac{{\rm det} [\Sigma^{-1}_{\w}(\X)]^{\frac{1}{2}}}{(2\pi \Phi_{\w}(\X))^{\frac{m}{2}}} e^{-\frac{q_{\w}(\Y,\X)}{2\Phi_{\w}(\X)}}$.

Finally, we can get the prediction model $\FF[\mu_{\w}(\X),\Sigma^{-1}_{\w}(\X),\Phi_{\w}(\X)]$, whose framework is illustrated in Figure~\ref{framework}. Individual and collaborative uncertainty are indirectly learned by the $\Sigma^{-1}_{\w}(\X)$.

\textbf{Loss function.}
On the basis of the square-root-free Cholesky decomposition and the parameterized form of $p(\Y|\X;\w)$, the Laplace collaborative uncertainty loss function is then:
\begin{eqnarray}
\vspace{-1mm}
\begin{small}
\LL_{\rm Lap-cu}(\w) = \frac{1}{2} \frac{1}{N}\sum\limits_{i=1}^{N}  [\frac{q_{\w}(\Y^i,\X^i)}{\Phi_{\w}(\X^i)}+m\log \Phi_{\w}(\X^i) - \sum\limits_{j=1}^{m} \log (d_{jj}^{i})]. 
\label{loss_la}
\end{small}
\vspace{-1mm}
\end{eqnarray}
where $d^{i}_{jj}$ is the $j$-th diagonal element in $D_{\w}(\X^{i})$. The parameters of $\mu_{\w}(\cdot)$, $\Sigma^{-1}_{\w}(\cdot)$ and $\Phi_{\w}(\cdot)$ are updated by minimizing (\ref{loss_la}). And $q_{\w}(\cdot,\cdot)$ is related to $\mu_{\w}(\cdot)$ and $\Sigma^{-1}_{\w}(\cdot)$, and $d^{i}_{jj}$ belongs to $\Sigma^{-1}_{\w}(\cdot)$.

\vspace{-1.5mm}
\subsection{Discussion}
\label{Discussion}
\label{3.4}
\vspace{-1.5mm}

After presenting how to quantify collaborative uncertainty in Section~\ref{3.2} and Section~\ref{3.3}, here we discuss the nature of collaborative uncertainty.

As mentioned in Section~\ref{3.2}, we can divide a prediction model for multi-agent trajectory forecasting into two types: individual models and collaborative models. An individual model predicts the future trajectory and the corresponding uncertainty for each agent independently; a collaborative model leverages an interaction module to explicitly capture the interactions among multiple agents, which makes all the predicted trajectories correlated. Moreover, this interaction modeling procedure can bring extra uncertainty to the model; in other words, we consider that the interaction modeling in a prediction model leads to collaborative uncertainty (CU).

\begin{figure}[tbh!]
\begin{center}

\centerline{
\includegraphics[width=1.\columnwidth]{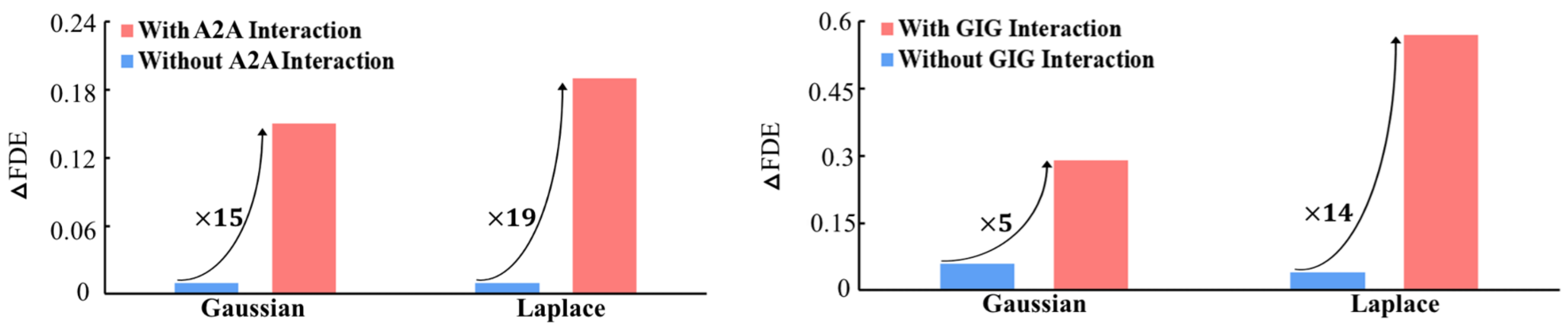}
 	}
\vspace{-.8em}
\caption{\textbf{CU modeling gains more when a prediction model includes interaction modeling.} Blue and red bars reflect the gains from CU modeling for a prediction model w/o an interaction module respectively. $\Delta$FDE is the difference of FDE values between versions w/o CU modeling.}
\label{compare_a2a}
\end{center}
\vskip -0.3in
\end{figure}

To validate this, we can empirically compare how much impact of collaborative uncertainty modeling would bring to an individual model and a collaborative model. In our experiment, we use two  cutting-edge multi-agent trajectory forecasting models, LaneGCN~\cite{liang2020learning} and VectorNet~\cite{Gao_2020_CVPR}.\footnote{To focus on agents' behaviors, we remove the map information and the related modules in this experiment.} In LaneGCN/VectorNet, an agent-to-agent (A2A) module/a global-interaction-graph (GIG) module explicitly models the interactions among multiple agents. As illustrated in Figure~\ref{compare_a2a}, when we remove the A2A/GIG module from LaneGCN/VectorNet, collaborative uncertainty modeling brings much less gain to LaneGCN/VectorNet; see Section~\ref{4.3} for more details of this experiment. This reflects that the cause of collaborative uncertainty mainly comes from the interaction modeling.

\vspace{-3mm}
\section{Experiments}
\label{experiment}
\vspace{-3mm}
We first use two self-generated synthetic datasets with a limited number of agents as the toy version of the real world problem. We use the simplified datasets to test our method's ability of capturing distribution information of the input data that obeys a certain type of multivariate distribution. We then conduct extensive experiments on two published benchmarks to prove the value of our proposed method in solving real world problems. We introduce the experiments in Sec. \ref{syn_data} and Sec. \ref{real_data}.

\vspace{-3mm}
\subsection{Toy Problem}
\label{syn_data}

\vspace{-2mm}

We define a toy problem to validate the capability of the proposed framework for accurate probability distribution estimation. The toy problem requires models to take the mutually correlated trajectories sampled from a given distribution as the input and output the mean and the covariance of the distribution.

As far as we know, in real-world datasets, we only have the ground truth for the predicted trajectory, which is the mean of the distribution while have no access to the ground truth of the uncertainty, which is the covariance matrix of the distribution. Therefore, we generate two synthetic datasets with the ground truth for both mean and covariance matrix given two distributions respectively.

\textbf{Datasets.} We generate two synthetic datasets for ternary Gaussian and ternary Laplace distribution respectively. Each dataset contains training, validation and test sets, which have 36000, 7000 and 7000 instances respectively. Each instance includes the trajectories of three agents, which consist of the two-dimensional point coordinates of the three agents at 50 different timestamp. In each instance, the trajectories of three agents are sampled from a ternary joint Gaussian/Laplace distribution. Generation details are provided in the appendix.

\textbf{Implementation details.} The network architecture used in the experiment contains an encoder and two decoders, all of which are four-layer MLPs. The neural network outputs the predicted mean and covariance matrix of the given ternary Gaussian/Laplace distribution. Although the ground truth covariance matrix is known for synthetic datasets, they are not used in training. We train the network using the previous uncertainty modeling method and our proposed method on each synthetic dataset.

\textbf{Metric.} Here, we adopt three metrics for evaluation: the $\ell_2$ distances between the estimated mean and the ground truth mean, the $\ell_1$ distances between the estimated covariance matrix and the ground truth covariance matrix and the KL divergence between the ground truth distribution and the estimated distribution. We provide metrics computing details in the appendix.

\begin{figure}[h]
\begin{center}
\centerline{\subfigure[Gaussian Synthetic Dataset]{
 		\includegraphics[scale=0.33]{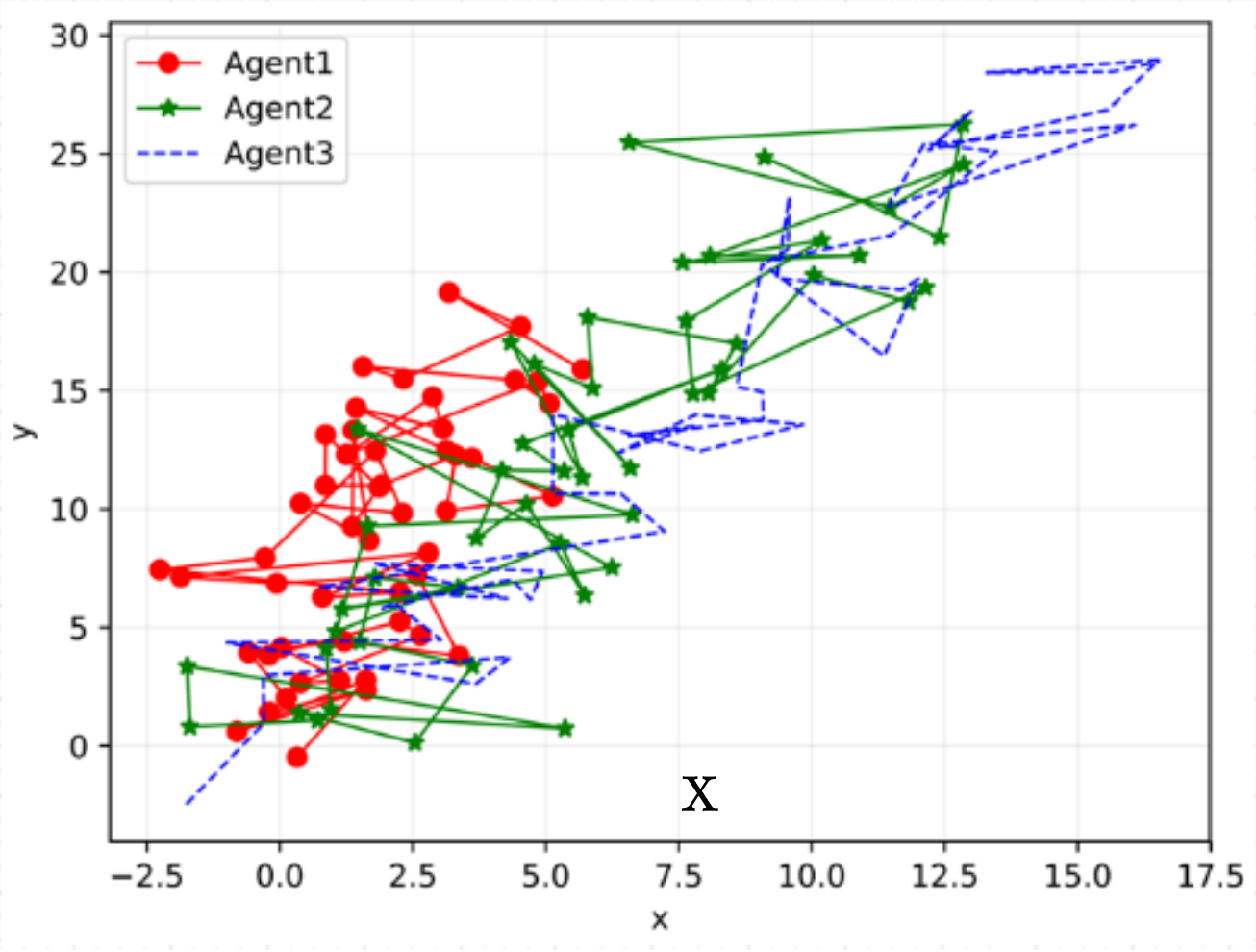}
 		\includegraphics[scale=0.33]{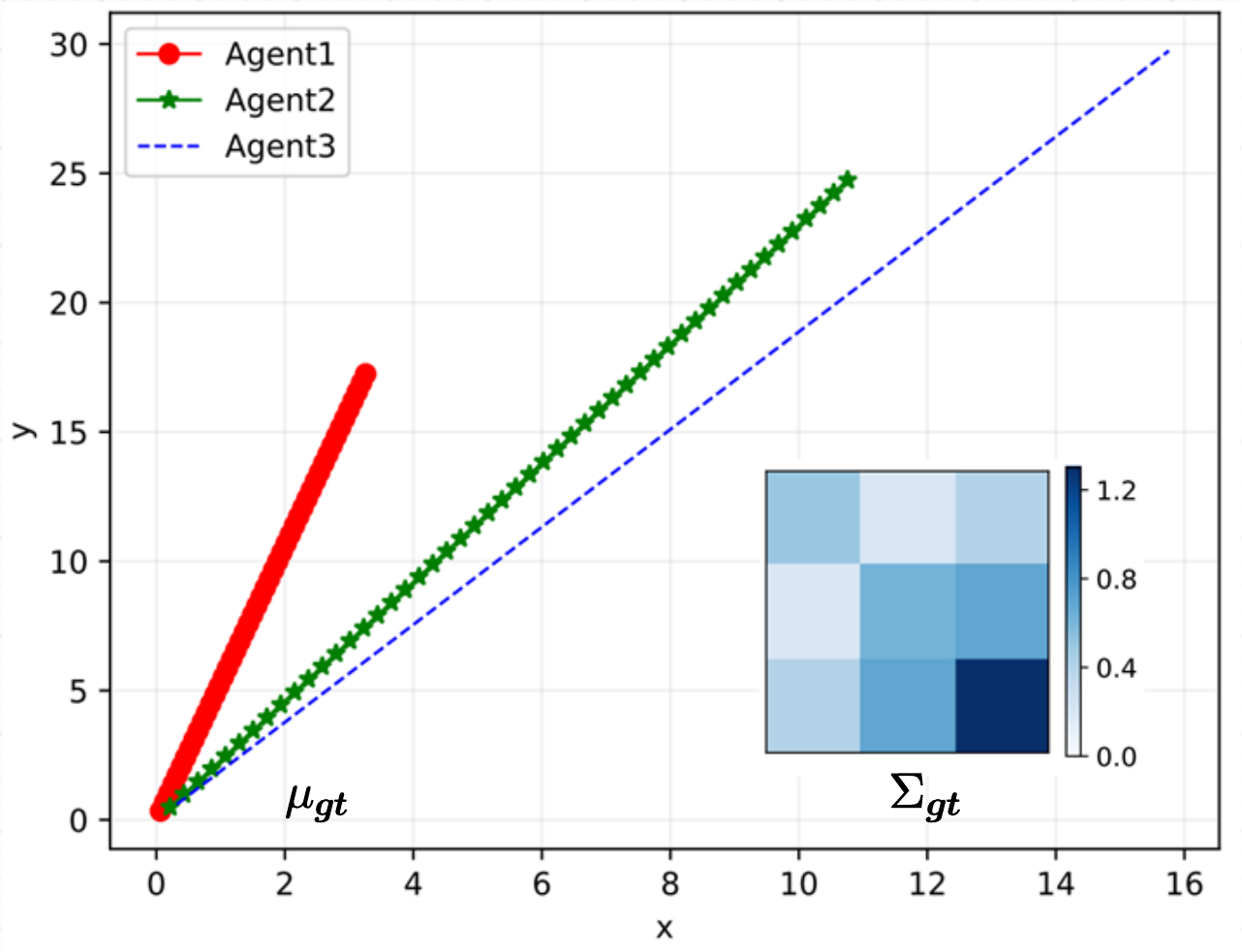}
		\includegraphics[scale=0.33]{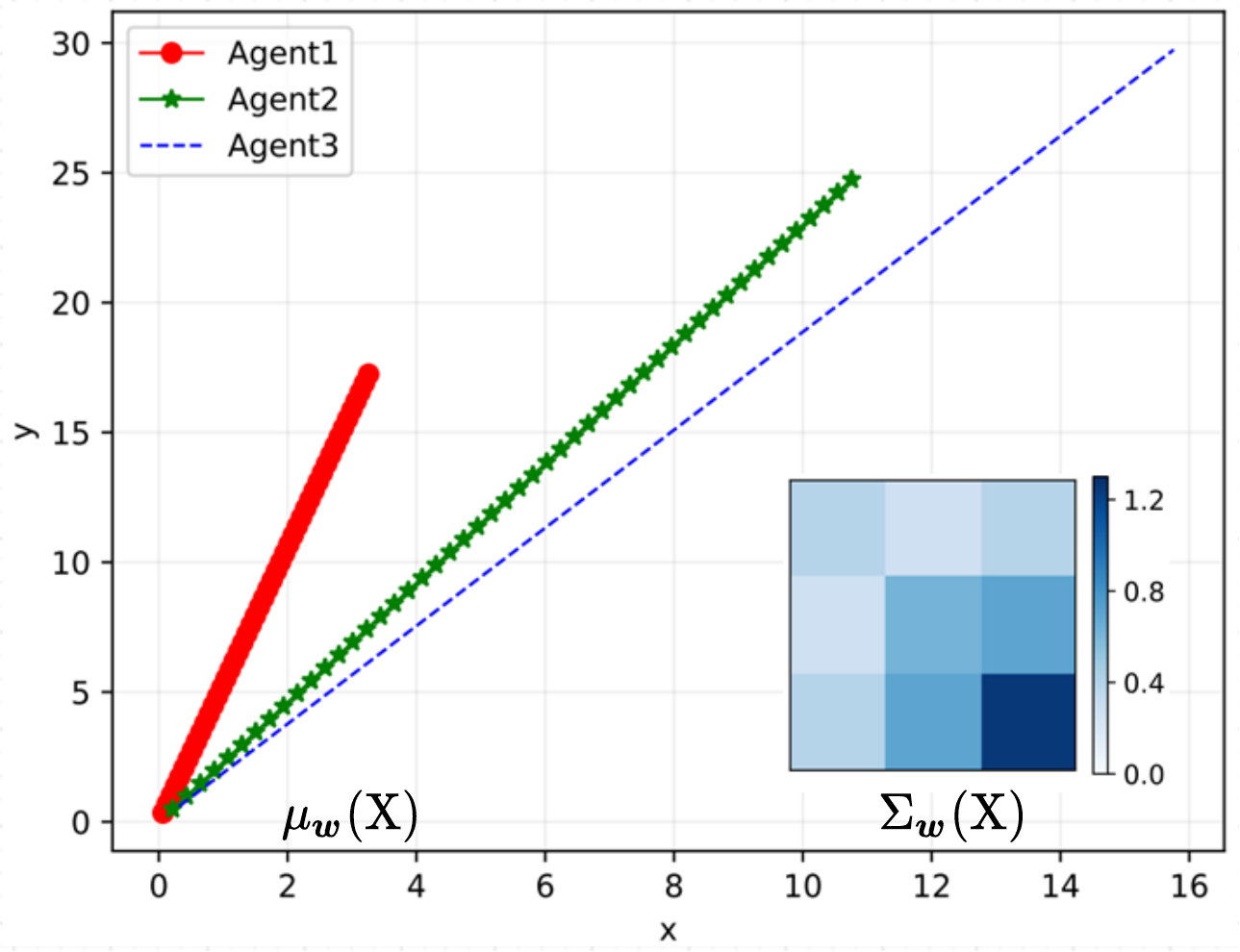}
 	}
 	}

\centerline{
 \subfigure[Laplace Synthetic Dataset]{
 		\includegraphics[scale=0.33]{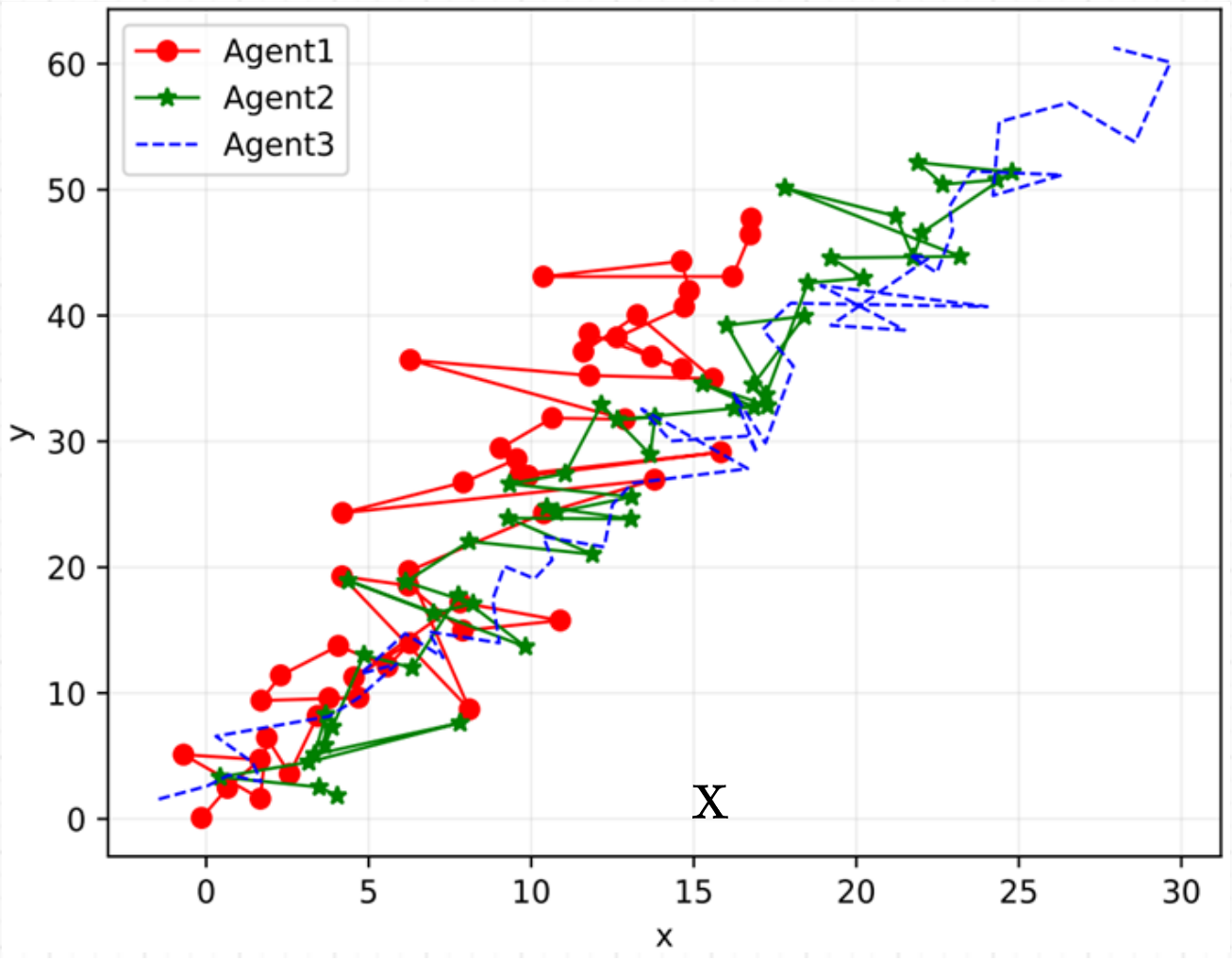}
 		\includegraphics[scale=0.33]{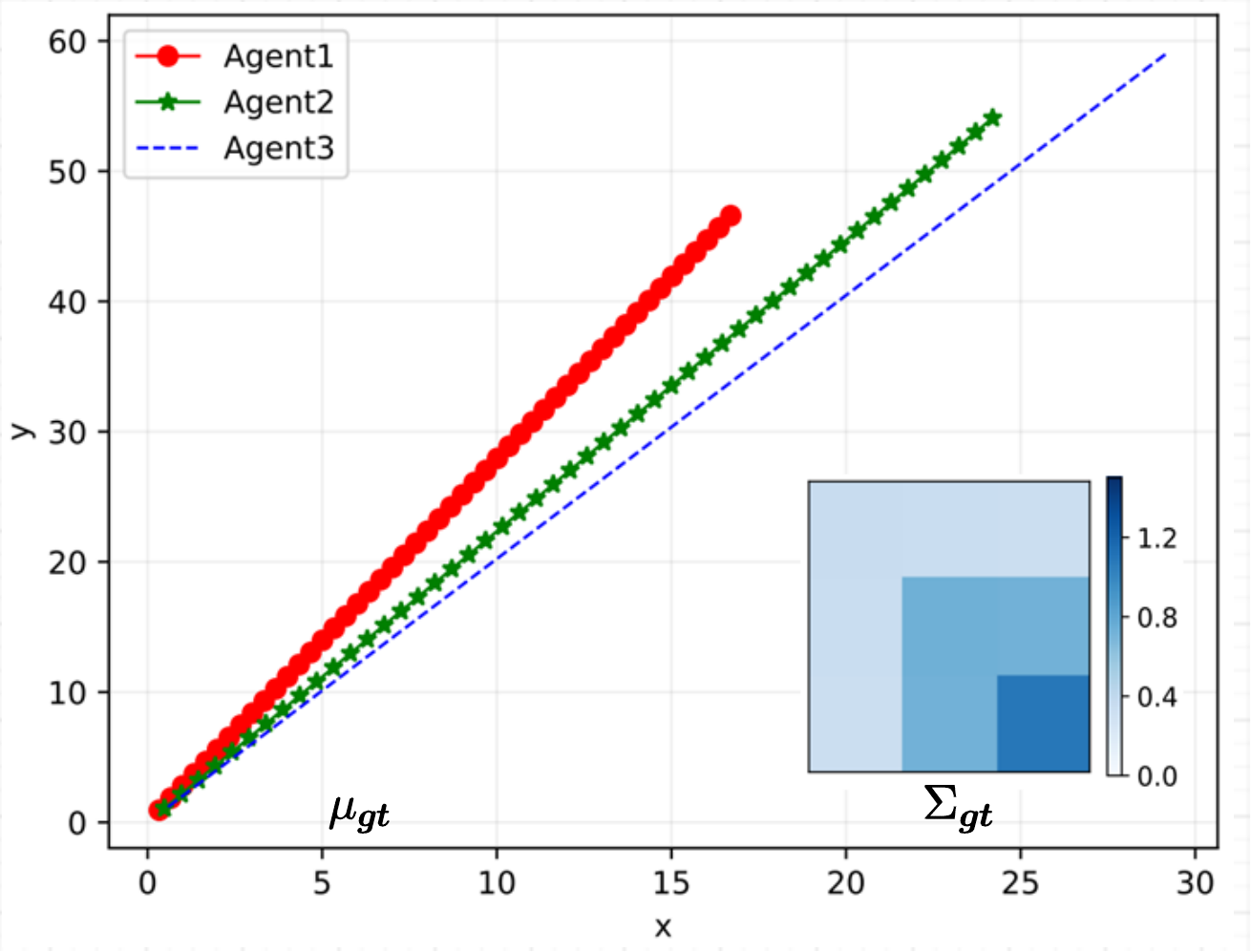}
		\includegraphics[scale=0.33]{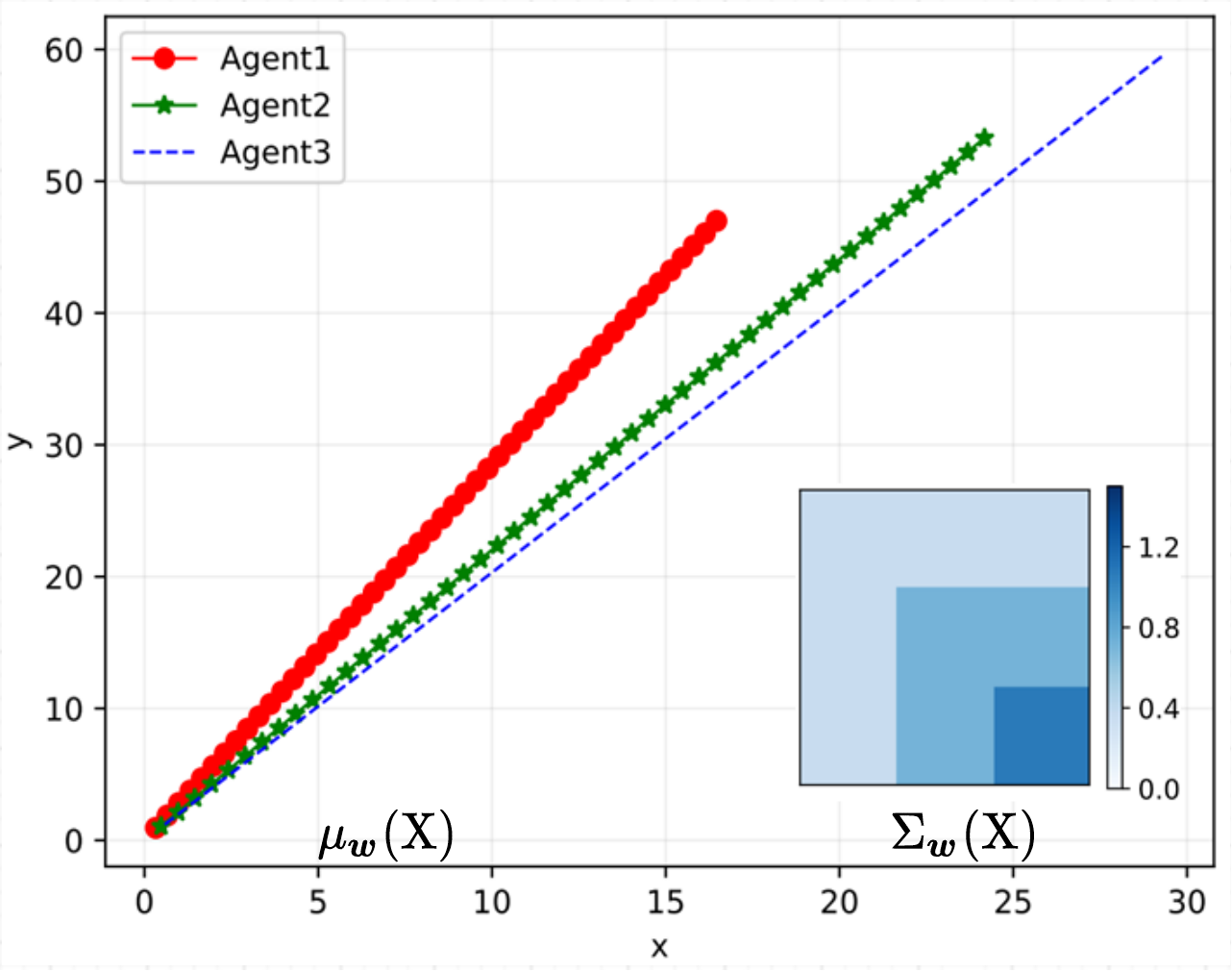}
 	}
	}
\caption{\textbf{Sample visualization on synthetic dataset.} Our proposed CU-based framework allows the model to learn the mean and covariance matrix of ground truth distribution accurately. Each input instance $\X$ is the numerical sum of a ground truth mean $\mu_{gt}$ and a random variable $\epsilon$ with the information of the ground truth covariance matrix $\Sigma_{gt}$. Please find more details in Appendix.}
\label{quali_sim}
\vspace{-2em}
\end{center}
\end{figure}

\textbf{Evaluation results.} The test set results of the synthetic dataset are shown in Table~\ref{simulation_r} and Figure~\ref{quali_sim}. Since the previous uncertainty modeling method only models IU (individual uncertainty) but our method models both IU and CU (collborative uncertainty), our method allows the model to estimate a more accurate mean and covariance matrix on a given distribution, which leads to a much lower KL divergence between the ground truth distribution and the estimated distribution. 

\begin{table}[tbh]
\begin{minipage}{.48\textwidth}
\begin{center}
\begin{footnotesize}
\begin{sc}
\setlength{\belowcaptionskip}{-0.15cm} 
\caption{Comparison with the prior uncertainty modeling method on synthetic datasets under the two different assumptions of distribution. $\mu_{\!\w}(\!\X\!)$: the estimated mean. $ \mu_{gt}$: the ground truth mean. $\Sigma_{\!\w}(\!\X\!)$: the estimated covariance matrix. $\Sigma_{gt}$: the ground truth covariance matrix. KL: the KL divergence $D_{KL}(p_{g}(\!\X\!)||p_{e}(\!\X\!))$, where $p_{e}(\!\X\!)$ is the estimated distribution and $p_{g}(\!\X\!)$ is the ground truth distribution.}\label{simulation_r}
\begin{tabular}{c|cc|cc}

\toprule
& \multicolumn{2}{c|}{\scriptsize Gaussian}&\multicolumn{2}{c}{\scriptsize Laplace}  \\ 
& {\scriptsize IU Only} & \multicolumn{1}{c|}{\scriptsize IU + CU} & {\scriptsize IU Only}  & {\scriptsize IU + CU} \\ \hline
{$\ell_{2}$ of $\mu$}
& 0.68  & \textbf{0.49}    & 0.42    & \textbf{0.34}  \\ 
{$\ell_1$ of $\Sigma$}
& 1.98  & \textbf{1.01}  & 1.96  & \textbf{1.13}  \\ 
KL 
& 6.68  & \textbf{0.40}  & 12.6  & \textbf{1.65}  \\
\bottomrule
\end{tabular}
\end{sc}
\end{footnotesize}
\end{center}
\end{minipage}%
\hspace{.03\textwidth}
\begin{minipage}{.48\textwidth}
\setlength{\belowcaptionskip}{-0.15cm}
\caption{Comparison with SOTA methods on Argoverse test set. CU boosts performances in single future prediction.}
\label{argo_test}
\vspace{-.5em}
\begin{center}
\begin{footnotesize}
\begin{sc}
\begin{tabular}{l|cc}
\toprule
Method & ADE
 & FDE
\\
\midrule
Argo Baseline (NN)~\cite{Chang_2019_CVPR} & 3.45 & 7.88
 \\
Argo Baseline~\cite{Chang_2019_CVPR} & 2.96 & 6.81
 \\
uulm-mrm~\cite{DBLP:conf/icra/CuiRCLNHSD19,DBLP:journals/corr/abs-1808-05819}& 1.90 & 4.19
 \\
VectorNet~\cite{Gao_2020_CVPR} & 1.81 & 4.01
 \\
TNT~\cite{DBLP:journals/corr/abs-2008-08294} & 1.78  & 3.91
\\
Jean~\cite{DBLP:conf/icra/MercatGZSBG20} & 1.74  & 4.24 
\\
LaneGCN~\cite{liang2020learning} & 1.71  & 3.78 
\\
  \hline
 \multicolumn{3}{l}{Ours:}
 \\
\hline
LaneGCN {\tiny(Our Implementation)} & 1.76
 & 3.84
 \\
Using $\LL_{\rm Gau-cu}$ & 1.73 & 3.83
 \\
Using $\LL_{\rm Lap-cu}$  & \textbf{1.70}  & \textbf{3.74}

 \\
 \bottomrule
\end{tabular}

\end{sc}
\end{footnotesize}
\end{center}
\end{minipage}
\end{table}
\vspace{-2em}

\vspace{4mm}
\subsection{Real World Problem}
\label{real_data}
\vspace{-2mm}

The experiments results above show the superiority of our method for simplified problems. Here we further validate its ability on two real-world large-scale autonomous driving datasets in single future prediction. We also conduct ablation studies in this subsection to present the effectiveness of the method design.

\vspace{-2mm}
\subsubsection{Experiment setup}
\textbf{Datasets.} \emph{Argoverse}~\cite{Chang_2019_CVPR} and \emph{nuScenes}~\cite{nuscenes2019} are two widely used multi-agent trajectory forecasting benchmarks. \emph{Argoverse} has over 30K scenarios collected in Pittsburgh and Miami. Each scenario is a sequence of frames sampled at 10 Hz. The sequences are split as training, validation and test sets, which have 205942, 39472 and 78143 sequences respectively. \emph{nuScenes} collects 1000 scenes in Boston and Singapore. Each scene is annotated at 2 Hz and is 20s long. The prediction instances are split as training, validation and test sets, which have 32186, 8560 and 9041 instances respectively. For Argoverse, we forecast future trajectories for 3s based on the observed trajectories of 2s. For nuScenes, we forecast future trajectories for 6s based on the observed trajectories of 2s.

\vspace{-0.02in}
\textbf{Metrics.} We adopt two extensively used multi-agent trajectory forecasting metrics, Average Displacement Error (ADE) and Final Displacement Error (FDE). ADE is defined as the average of pointwise $\ell_2$ distances between the predicted trajectory and ground truth. FDE is defined as the $\ell_2$ distance between the final points of the prediction and ground truth (ADE and FDE are measured in meters). 

\vspace{-0.02in}
\textbf{Implementation details.} The proposed model is implemented follow the basis of LaneGCN~\cite{liang2020learning} and VectorNet~\cite{Gao_2020_CVPR}, two cutting-edge multi-agent trajectory forecasting models. We implement the encoder of LaneGCN/VectorNet as the trajectory encoder in our proposed model. We further use the four-layer multilayer perceptrons (MLPs) to respectively implement three decoders $\mu_{\w}(\cdot)$, $\Sigma_{\w}(\cdot)$ and $\Phi_{\w}(\cdot)$. Note that all experiments are based on our own implementation of LaneGCN/VectorNet, which may not perform as well as the official LaneGCN/VectorNet.
\vspace{-1.5mm}

\begin{table*}[tbh]
\caption{Ablation on covariance $\Sigma$ with chosen probability density function (\textbf{PDF}) and Interaction module (\textbf{INT.}). \textbf{DIA} denotes the diagonal matrix (individual uncertainty). \textbf{FULL} denotes the full matrix  (individual and collaborative uncertainty). On Argoverse and nuScenes, both LaneGCN and VectorNet with individual and collaborative uncertainty surpasses the ones with individual uncertainty only. Collaborative uncertainty makes a larger impact for the model with an interaction module.}
\vspace{-2mm}
\label{ablation_un}
\vskip -0.4in
\begin{center}
\begin{footnotesize}
\begin{sc}
\begin{tabular}{c|c|c|c|cc|c|cc|c}
\toprule
\multirow{2}{*}{Method} &\multirow{2}{*}{Dataset} & \multirow{2}{*}{PDF Type}&\multirow{2}{*}{Int.} & \multicolumn{2}{c|}{ADE} & \multirow{2}{*}{$\Delta$ADE} & \multicolumn{2}{c|}{FDE} & \multirow{2}{*}{$\Delta$FDE} 
\\
&  & & &DIA & FULL& & DIA & FULL &\\
\hline
\multirow{8}{*}{\scriptsize LaneGCN}&\multirow{4}{*}{\scriptsize Argoverse}&\multirow{2}{*}{\scriptsize Gaussian}&  $\times$ & 1.69 & 1.67 & 0.02 & 3.88 & 3.85 & 0.03\\
&& &  $\surd$ & 1.45 & 1.42 & \textbf{0.03} & 3.19 & 3.14 & \textbf{0.05}
\\
\cline{3-10}
&&\multirow{2}{*}{\scriptsize Laplace }&  $\times$ & 1.67 & 1.67 & 0.00 & 3.82 & 3.82 & 0.00
\\
&&&$\surd$ & 1.43 & \textbf{1.41} & \textbf{0.02} & 3.16 & \textbf{3.11} & \textbf{0.05}
\\
\cline{2-10}
&\multirow{4}{*}{\scriptsize nuScenes}&\multirow{2}{*}{\scriptsize Gaussian}&$\times$ & 4.60 & 4.58 & 0.02 & 11.02 & 11.01 & 0.01
\\
&&& $\surd$ & 4.47 & 4.39 & \textbf{0.08} & 10.59 & 10.44 & \textbf{0.15}
\\
\cline{3-10}
&&\multirow{2}{*}{\scriptsize Laplace }& $\times$ & 4.53 & 4.52 & 0.01 & 10.93 & 10.92 & 0.01
\\
&&& $\surd$ & 4.34 & \textbf{4.25} & \textbf{0.09} & 10.34 & \textbf{10.15} & \textbf{0.19}
\\
\hline
\hline
\multirow{8}{*}{\scriptsize VectorNet}&\multirow{4}{*}{\scriptsize Argoverse}&\multirow{2}{*}{\scriptsize Gaussian}& $\times$ & 1.82 & 1.78 & 0.04 & 4.16 & 4.08 & 0.08
\\
&&& $\surd$ & 1.63 & 1.57 & \textbf{0.06} & 3.60 & 3.46 & \textbf{0.14}
\\
\cline{3-10}
&&\multirow{2}{*}{\scriptsize Laplace }& $\times$ & 1.78 & 1.76 & 0.02 & 4.06 & 4.02 & 0.04
\\
&&& $\surd$ & 1.56 & \textbf{1.52} & \textbf{0.04} & 3.42 & \textbf{3.34} & \textbf{0.08}
\\

\cline{2-10}
&\multirow{4}{*}{\scriptsize nuScenes}&\multirow{2}{*}{\scriptsize Gaussian}& $\times$ & 4.25 & 4.23 & 0.02 & 10.35 & 10.29
& 0.06 
\\
&&&$\surd$ & 4.07 & 3.99 & \textbf{0.08} & 9.86 & 9.57 & \textbf{0.29}
\\
\cline{3-10}
&&\multirow{2}{*}{\scriptsize Laplace}& $\times$ & 4.19 & 4.18 & 0.01 & 10.25 & 10.21 & 0.04
\\
&&& $\surd$ & 4.02 & \textbf{3.81} & \textbf{0.21} & 9.79 & \textbf{9.22} & \textbf{0.57}
\\
\bottomrule
\end{tabular}
\end{sc}
\end{footnotesize}
\end{center}
\vskip -0.2in
\end{table*}
\label{as_A2A}

\subsubsection{Results}
\label{4.2}

\textbf{Evaluation results on benchmark datasets.} In this subsection, we implement our proposed framework based on LaneGCN~\cite{liang2020learning} as it is the SOTA model in multi-agent trajectory forecasting. We compare our proposed methods in the Argoverse trajectory forecasting benchmark with two official baselines, including Argoverse Baseline~\cite{Chang_2019_CVPR} and Argoverse Baseline (NN)~\cite{Chang_2019_CVPR}, and five SOTA methods of this benchmark: LaneGCN~\cite{liang2020learning}, TNT~\cite{DBLP:journals/corr/abs-2008-08294}, Jean~\cite{DBLP:conf/icra/MercatGZSBG20}, VectorNet~\cite{Gao_2020_CVPR} and uulm-mrm~\cite{DBLP:conf/icra/CuiRCLNHSD19,DBLP:journals/corr/abs-1808-05819}. Table~\ref{argo_test} shows that although our implementation of LaneGCN appears less good than the official implementation of LaneGCN, our implementation of LaneGCN with the individual and collaborative uncertainty notably outperforms all of the other competing methods in both ADE and FDE. Therefore, our proposed collaborative uncertainty modeling enhances the SOTA prediction models.
\begin{figure}[tbh!]
\begin{center}
\centerline{\subfigure[Scenario \rom{1}]{
         \label{fig:s1}
 		\includegraphics[scale=0.23]{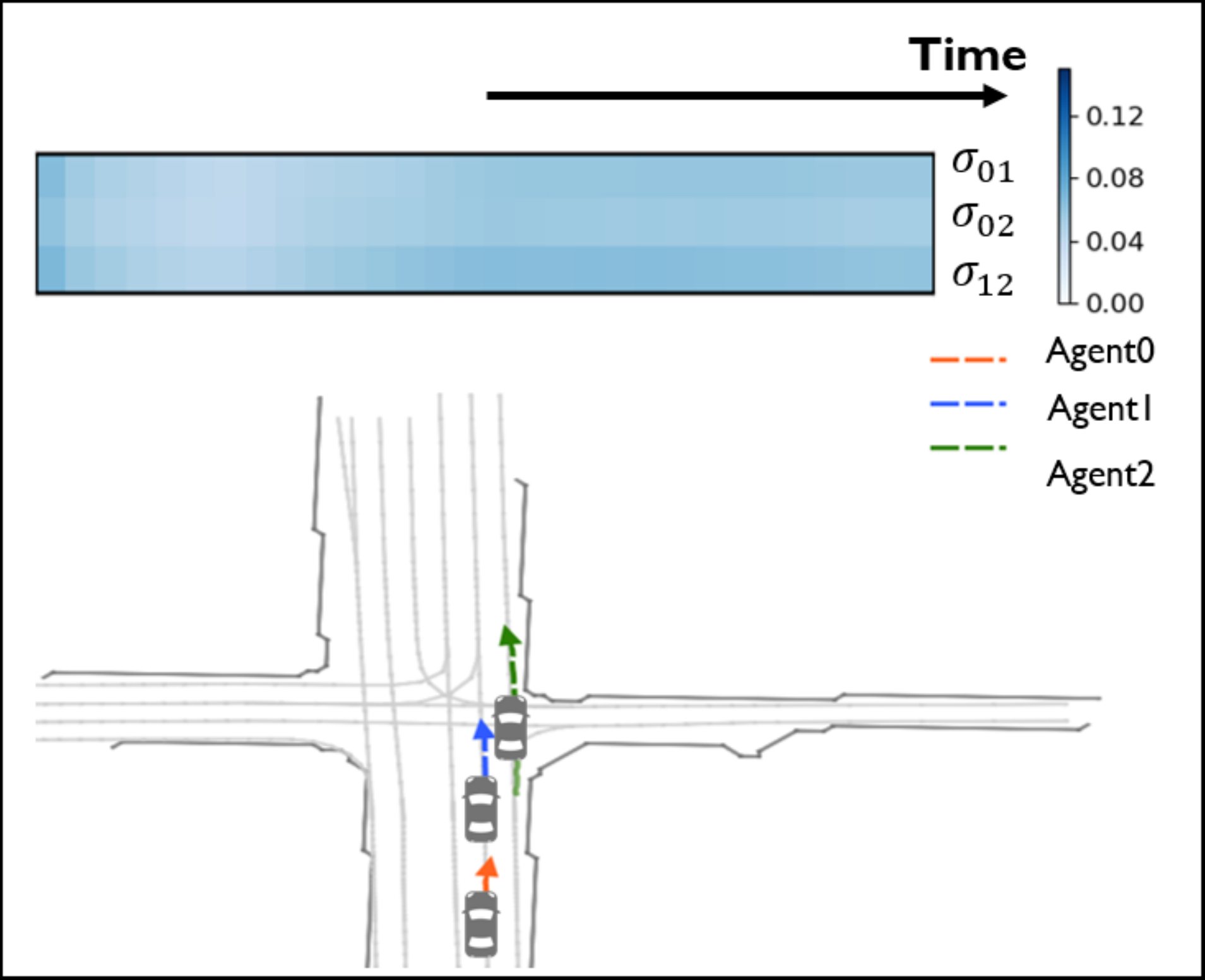}
 	}
 \subfigure[Scenario \rom{2}]{
         \label{fig:s2}
 		\includegraphics[scale=0.23]{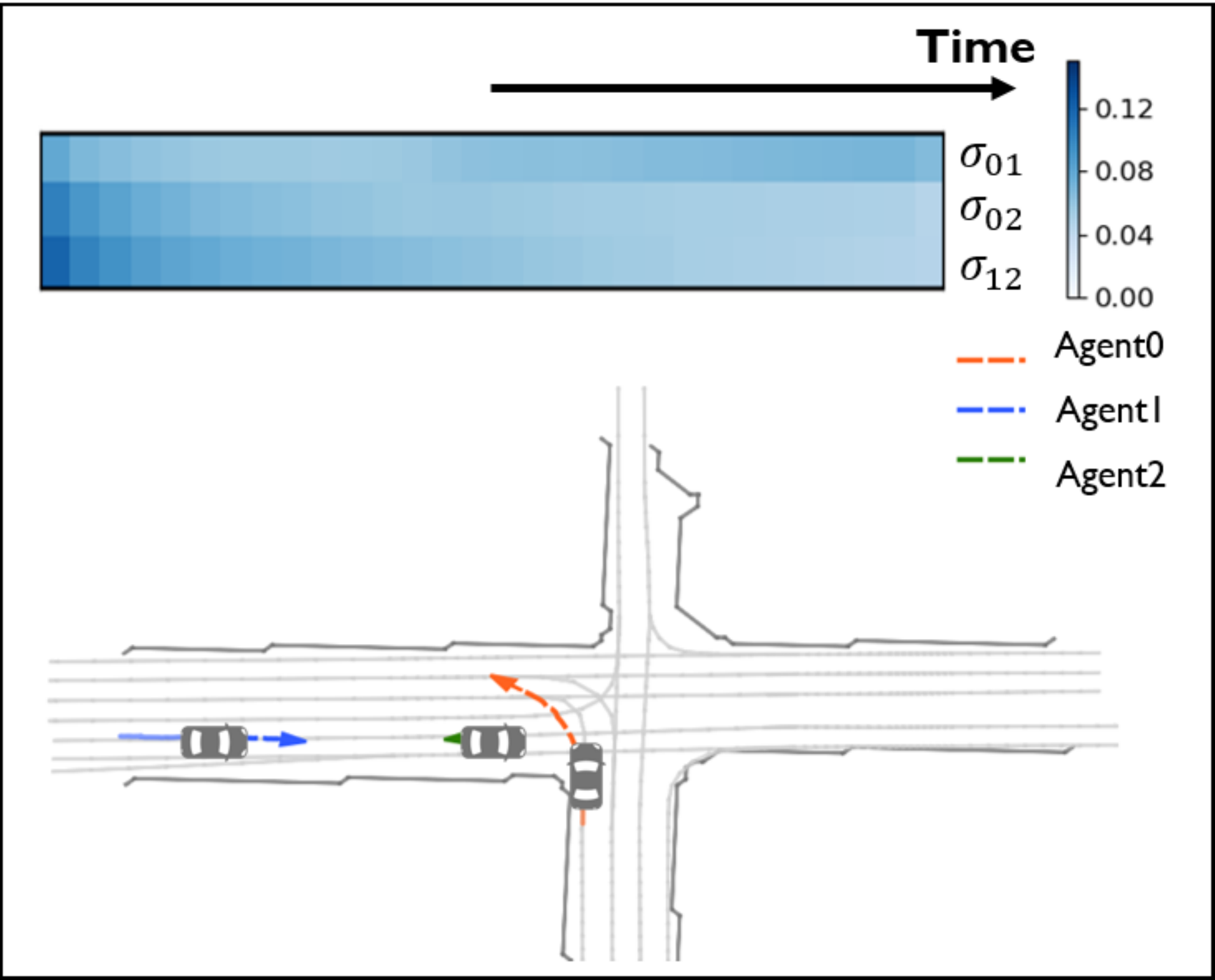}
 	}
 \subfigure[Scenario \rom{3}]{
         \label{fig:s3}
		\includegraphics[scale=0.23]{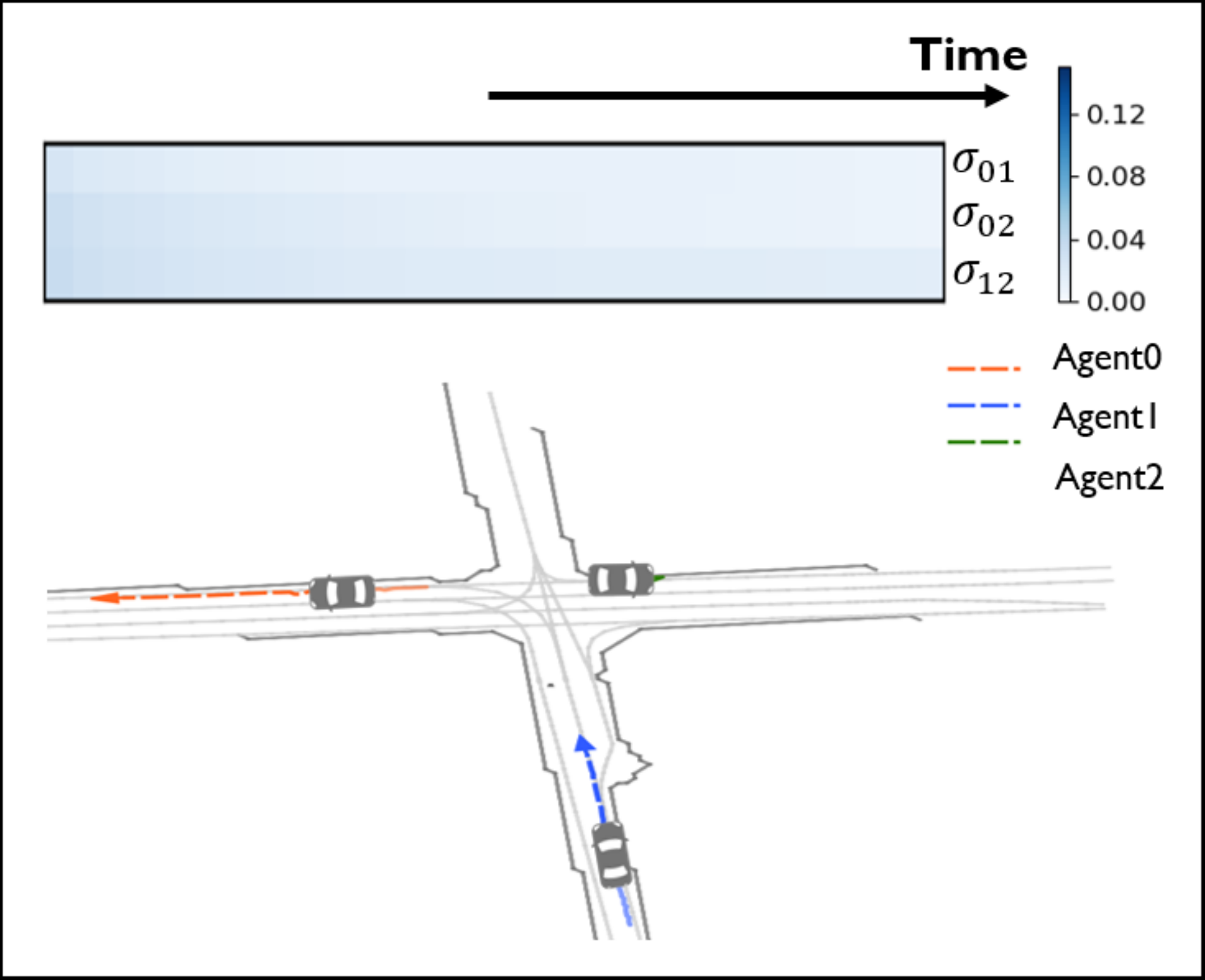}
	}	
	}
\caption{\textbf{Visualization of CU on Argoverse dataset}. (a) In scenario \rom{1}, Agent0 and Agent1 are driving on the same road, which is next to the road where Agent2 is driving. This type of scenario might generate complicated interactive information making $\sigma_{01}$, $\sigma_{02}$ and $\sigma_{12}$ show a non-monotonic change over time. (b) In scenario \rom{2}, Agent0 and Agent1 are moving towards each other, thus $\sigma_{01}$ increases over time. Agent2 is parking at an intersection waiting for green light, as little new interactive information between Agent2 and the other two agents would be generated before the red light turns green, $\sigma_{02}$ and $\sigma_{12}$ decrease over time. (c) In scenario \rom{3}, Agent0, Agent1 and Agent2 are located in completely different areas heading to different directions, $\sigma_{01}$, $\sigma_{02}$ and $\sigma_{12}$ are close to zero.}
\label{Visualization}
\end{center}
\end{figure}
\vspace{-1.7em}

\textbf{Visualization of collaborative uncertainty.}\footnote{These visualization results are based on Gaussian CU. Please see more visualization results in the appendix.} To visually understand which factor might influence the value of collaborative uncertainty in multi-agent trajectory forecasting, we present the visualization results generated by our model in Figure~\ref{Visualization}. We visualize 3 scenarios, and each scenario includes 3 agents (orange, blue and green lines) trajectories (solid lines are the past trajectories and dashed lines are the future trajectories) and their corresponding collaborative uncertainty values changing over the last 30 frames (the heatmap, where $\sigma_{ij}$ denotes the collaborative uncertainty of agent $i$ and agent $j$). These results show that the value of collaborative uncertainty is highly related to the amount of the interactive information among agents.

\subsubsection{Ablation study}
\label{4.3}

We study: 1) how different approaches of uncertainty modeling would affect the prediction model; 2) how the interaction module would influence the effectiveness of collaborative uncertainty modeling. In this part, we adopt LaneGCN/VectorNet as our trajectory encoder for proving that our proposed method can be used as a plug-in module to improve the performances of existing models in multi-agent trajectory forecasting. To focus on the agents' behaviors, we ignore map information and map-related modules in LaneGCN/VectorNet; we only use the agent encoder of LaneGCN/VectorNet, which extracts intermediate feature from agents' past trajectories and actor-to-actor (A2A) module/global-interaction-graph (GIG) module that exploits the interactions among agents. In this part, the experiments are conducted on the validate sets of Argoverse and nuScenes benchmarks.

\textbf{Different approaches of uncertainty modeling.} We consider two approaches of uncertainty modeling: assuming the covariance matrix is a diagonal matrix (DIA) to model individual uncertainty only, and assuming the covariance matrix is a full matrix (FULL) to model both individual and collaborative uncertainty. Based on the results in Table~\ref{ablation_un}, we see that 1) modeling individual and collaborative uncertainty together (FULL) is consistently better than modeling individual uncertainty only (DIA); and 2) our proposed Laplace CU-based framework enables LaneGCN and VectorNet to achieve the best performances on ADE \& FDE metrics on both Argoverse and nuScenes benchmarks. These results reflect that our proposed collaborative uncertainty modeling can work as a plugin module to significantly improve the prediction performance of existing models.

\textbf{Effects of interaction module.} A2A/GIG module is the interaction module in LaneGCN/VectorNet. We study the effects of the interaction module in collaborative uncertainty modeling via the ablation study on A2A/GIG module. $\Delta$ADE/FDE is the difference of ADE/FDE values between versions with and without collaborative uncertainty. Higher values reflect bigger gains brought by modeling collaborative uncertainty. From Table 4, gains from modeling collaborative uncertainty in a collaborative model (with A2A/GIG) are much greater than in an individual model (without A2A/GIG). Figure~\ref{compare_a2a} visualizes the $\Delta$FDE on nuScenes benchmark; see discussions in Section~\ref{3.4}.

\vspace{-0.1in}
\section{Conclusions}
\label{conclusions}
\vspace{-3mm}
This work proposes a novel probabilistic collaborative uncertainty (CU)-based framework for multi-agent trajectory forecasting. Its key novelty is conceptualizing and modeling CU introduced by interaction modeling. The experimental results in this work demonstrate the ability of the proposed CU-based framework to boost the performance of SOTA prediction systems, especially collaborative-model-based systems. The proposed framework potentially lead to more reliable self-driving systems.

This work shows the promise of our method for single-future trajectory prediction, which predicts the single best trajectory for agents. With growing research interests in predicting multiple potential trajectories for agents (i.e. multi-future trajectory prediction), we leave the generalization of our method to multi-future trajectory prediction as an important direction for future research.

\section*{Acknowledgement}
This work was supported in part by the National Key R$\&$D Program of China under Grant 2021YFB1714800, the National Natural Science Foundation of China under Grant 6217010074, 62173034, and the Science and Technology Commission of Shanghai Municipal under Grant 21511100900.

{\small
\bibliographystyle{unsrt}
\bibliography{main}

\begin{thebibliography}{10}

\bibitem{DBLP:journals/thms/StahlDJ14}
Patrick Stahl, Birsen Donmez, and Greg~A. Jamieson.
\newblock Anticipation in driving: The role of experience in the efficacy of
  pre-event conflict cues.
\newblock {\em {IEEE} Trans. Hum. Mach. Syst.}, 44(5):603--613, 2014.

\bibitem{2016Supporting}
Patrick Stahl, Birsen Donmez, and Greg~A Jamieson.
\newblock Supporting anticipation in driving through attentional and
  interpretational in-vehicle displays.
\newblock {\em Accident Analysis and Prevention}, 91(jun.):103--113, 2016.

\bibitem{liang2020learning}
Ming Liang, Bin Yang, Rui Hu, Yun Chen, Renjie Liao, Song Feng, and Raquel
  Urtasun.
\newblock Learning lane graph representations for motion forecasting.
\newblock In {\em ECCV}, 2020.

\bibitem{Gao_2020_CVPR}
Jiyang Gao, Chen Sun, Hang Zhao, Yi~Shen, Dragomir Anguelov, Congcong Li, and
  Cordelia Schmid.
\newblock Vectornet: Encoding hd maps and agent dynamics from vectorized
  representation.
\newblock In {\em Proceedings of the IEEE/CVF Conference on Computer Vision and
  Pattern Recognition (CVPR)}, June 2020.

\bibitem{Ye_2021_CVPR}
Maosheng Ye, Tongyi Cao, and Qifeng Chen.
\newblock Tpcn: Temporal point cloud networks for motion forecasting.
\newblock In {\em Proceedings of the IEEE/CVF Conference on Computer Vision and
  Pattern Recognition (CVPR)}, pages 11318--11327, June 2021.

\bibitem{gilles2021home}
Thomas Gilles, Stefano Sabatini, Dzmitry Tsishkou, Bogdan Stanciulescu, and
  Fabien Moutarde.
\newblock Home: Heatmap output for future motion estimation, 2021.

\bibitem{DBLP:conf/icc/XiaoZHY19}
Ke~Xiao, Jianyu Zhao, Yunhua He, and Shui Yu.
\newblock Trajectory prediction of {UAV} in smart city using recurrent neural
  networks.
\newblock In {\em 2019 {IEEE} International Conference on Communications, {ICC}
  2019, Shanghai, China, May 20-24, 2019}, pages 1--6. {IEEE}, 2019.

\bibitem{DBLP:conf/icml/JetchevT09}
Nikolay Jetchev and Marc Toussaint.
\newblock Trajectory prediction: learning to map situations to robot
  trajectories.
\newblock In Andrea~Pohoreckyj Danyluk, L{\'{e}}on Bottou, and Michael~L.
  Littman, editors, {\em Proceedings of the 26th Annual International
  Conference on Machine Learning, {ICML} 2009, Montreal, Quebec, Canada, June
  14-18, 2009}, volume 382 of {\em {ACM} International Conference Proceeding
  Series}, pages 449--456. {ACM}, 2009.

\bibitem{DBLP:conf/aimech/RosmannO0B17}
Christoph R{\"{o}}smann, Malte Oeljeklaus, Frank Hoffmann, and Torsten Bertram.
\newblock Online trajectory prediction and planning for social robot
  navigation.
\newblock In {\em {IEEE} International Conference on Advanced Intelligent
  Mechatronics, {AIM} 2017, Munich, Germany, July 3-7, 2017}, pages 1255--1260.
  {IEEE}, 2017.

\bibitem{Zhao_2019_CVPR}
Tianyang Zhao, Yifei Xu, Mathew Monfort, Wongun Choi, Chris Baker, Yibiao Zhao,
  Yizhou Wang, and Ying~Nian Wu.
\newblock Multi-agent tensor fusion for contextual trajectory prediction.
\newblock In {\em Proceedings of the IEEE/CVF Conference on Computer Vision and
  Pattern Recognition (CVPR)}, June 2019.

\bibitem{Choi_2019_ICCV}
Chiho Choi and Behzad Dariush.
\newblock Looking to relations for future trajectory forecast.
\newblock In {\em Proceedings of the IEEE/CVF International Conference on
  Computer Vision (ICCV)}, October 2019.

\bibitem{salzmann2020trajectron++}
Tim Salzmann, Boris Ivanovic, Punarjay Chakravarty, and Marco Pavone.
\newblock Trajectron++: Dynamically-feasible trajectory forecasting with
  heterogeneous data.
\newblock {\em arXiv preprint arXiv:2001.03093}, 2020.

\bibitem{zeng2021lanercnn}
Wenyuan Zeng, Ming Liang, Renjie Liao, and Raquel Urtasun.
\newblock Lanercnn: Distributed representations for graph-centric motion
  forecasting.
\newblock {\em arXiv preprint arXiv:2101.06653}, 2021.

\bibitem{li2020evolvegraph}
Jiachen Li, Fan Yang, Masayoshi Tomizuka, and Chiho Choi.
\newblock Evolvegraph: Multi-agent trajectory prediction with dynamic
  relational reasoning.
\newblock {\em Proceedings of the Neural Information Processing Systems
  (NeurIPS)}, 2020.

\bibitem{DBLP:conf/nips/KosarajuSM0RS19}
Vineet Kosaraju, Amir Sadeghian, Roberto Mart{\'{\i}}n{-}Mart{\'{\i}}n, Ian~D.
  Reid, Hamid Rezatofighi, and Silvio Savarese.
\newblock Social-bigat: Multimodal trajectory forecasting using bicycle-gan and
  graph attention networks.
\newblock In Hanna~M. Wallach, Hugo Larochelle, Alina Beygelzimer, Florence
  d'Alch{\'{e}}{-}Buc, Emily~B. Fox, and Roman Garnett, editors, {\em Advances
  in Neural Information Processing Systems 32: Annual Conference on Neural
  Information Processing Systems 2019, NeurIPS 2019, December 8-14, 2019,
  Vancouver, BC, Canada}, pages 137--146, 2019.

\bibitem{Gal2016Uncertainty}
Yarin Gal.
\newblock {\em Uncertainty in Deep Learning}.
\newblock PhD thesis, University of Cambridge, 2016.

\bibitem{NIPS2017_2650d608}
Alex Kendall and Yarin Gal.
\newblock What uncertainties do we need in bayesian deep learning for computer
  vision?
\newblock In I.~Guyon, U.~V. Luxburg, S.~Bengio, H.~Wallach, R.~Fergus,
  S.~Vishwanathan, and R.~Garnett, editors, {\em Advances in Neural Information
  Processing Systems}, volume~30, pages 5574--5584. Curran Associates, Inc.,
  2017.

\bibitem{zhao2020uncertainty}
Xujiang Zhao, Feng Chen, Shu Hu, and Jin-Hee Cho.
\newblock Uncertainty aware semi-supervised learning on graph data, 2020.

\bibitem{der2009aleatory}
Armen Der~Kiureghian and Ove Ditlevsen.
\newblock Aleatory or epistemic? does it matter?
\newblock {\em Structural safety}, 31(2):105--112, 2009.

\bibitem{DBLP:conf/cvpr/BhattacharyyaFS18}
Apratim Bhattacharyya, Mario Fritz, and Bernt Schiele.
\newblock Long-term on-board prediction of people in traffic scenes under
  uncertainty.
\newblock In {\em 2018 {IEEE} Conference on Computer Vision and Pattern
  Recognition, {CVPR} 2018, Salt Lake City, UT, USA, June 18-22, 2018}, pages
  4194--4202. {IEEE} Computer Society, 2018.

\bibitem{DBLP:conf/corl/Jain0LXFSU19}
Ajay Jain, Sergio Casas, Renjie Liao, Yuwen Xiong, Song Feng, Sean Segal, and
  Raquel Urtasun.
\newblock Discrete residual flow for probabilistic pedestrian behavior
  prediction.
\newblock In Leslie~Pack Kaelbling, Danica Kragic, and Komei Sugiura, editors,
  {\em 3rd Annual Conference on Robot Learning, CoRL 2019, Osaka, Japan,
  October 30 - November 1, 2019, Proceedings}, volume 100 of {\em Proceedings
  of Machine Learning Research}, pages 407--419. {PMLR}, 2019.

\bibitem{DBLP:conf/cvpr/HongSP19}
Joey Hong, Benjamin Sapp, and James Philbin.
\newblock Rules of the road: Predicting driving behavior with a convolutional
  model of semantic interactions.
\newblock In {\em {IEEE} Conference on Computer Vision and Pattern Recognition,
  {CVPR} 2019, Long Beach, CA, USA, June 16-20, 2019}, pages 8454--8462.
  Computer Vision Foundation / {IEEE}, 2019.

\bibitem{DBLP:conf/cvpr/KendallGC18}
Alex Kendall, Yarin Gal, and Roberto Cipolla.
\newblock Multi-task learning using uncertainty to weigh losses for scene
  geometry and semantics.
\newblock In {\em 2018 {IEEE} Conference on Computer Vision and Pattern
  Recognition, {CVPR} 2018, Salt Lake City, UT, USA, June 18-22, 2018}, pages
  7482--7491. {IEEE} Computer Society, 2018.

\bibitem{ayhan2018test}
Murat~Seckin Ayhan and Philipp Berens.
\newblock Test-time data augmentation for estimation of heteroscedastic
  aleatoric uncertainty in deep neural networks.
\newblock 2018.

\bibitem{pmlr-v80-depeweg18a}
Stefan Depeweg, Jose-Miguel Hernandez-Lobato, Finale Doshi-Velez, and Steffen
  Udluft.
\newblock Decomposition of uncertainty in {B}ayesian deep learning for
  efficient and risk-sensitive learning.
\newblock In Jennifer Dy and Andreas Krause, editors, {\em Proceedings of the
  35th International Conference on Machine Learning}, volume~80 of {\em
  Proceedings of Machine Learning Research}, pages 1184--1193,
  Stockholmsmässan, Stockholm Sweden, 10--15 Jul 2018. PMLR.

\bibitem{feng2018towards}
Di~Feng, Lars Rosenbaum, and Klaus Dietmayer.
\newblock Towards safe autonomous driving: Capture uncertainty in the deep
  neural network for lidar 3d vehicle detection.
\newblock In {\em 2018 21st International Conference on Intelligent
  Transportation Systems (ITSC)}, pages 3266--3273. IEEE, 2018.

\bibitem{wang2019aleatoric}
Guotai Wang, Wenqi Li, Michael Aertsen, Jan Deprest, S{\'e}bastien Ourselin,
  and Tom Vercauteren.
\newblock Aleatoric uncertainty estimation with test-time augmentation for
  medical image segmentation with convolutional neural networks.
\newblock {\em Neurocomputing}, 338:34--45, 2019.

\bibitem{pmlr-v119-kong20b}
Lingkai Kong, Jimeng Sun, and Chao Zhang.
\newblock {SDE}-net: Equipping deep neural networks with uncertainty estimates.
\newblock In Hal~Daumé III and Aarti Singh, editors, {\em Proceedings of the
  37th International Conference on Machine Learning}, volume 119 of {\em
  Proceedings of Machine Learning Research}, pages 5405--5415. PMLR, 13--18 Jul
  2020.

\bibitem{8578672}
G.~{Dorta}, S.~{Vicente}, L.~{Agapito}, N.~D.~F. {Campbell}, and I.~{Simpson}.
\newblock Structured uncertainty prediction networks.
\newblock In {\em 2018 IEEE/CVF Conference on Computer Vision and Pattern
  Recognition}, pages 5477--5485, 2018.

\bibitem{DBLP:conf/nips/MonteiroFCPMKWG20}
Miguel Monteiro, Lo{\"{\i}}c~Le Folgoc, Daniel~Coelho de~Castro, Nick
  Pawlowski, Bernardo Marques, Konstantinos Kamnitsas, Mark van~der Wilk, and
  Ben Glocker.
\newblock Stochastic segmentation networks: Modelling spatially correlated
  aleatoric uncertainty.
\newblock In Hugo Larochelle, Marc'Aurelio Ranzato, Raia Hadsell,
  Maria{-}Florina Balcan, and Hsuan{-}Tien Lin, editors, {\em Advances in
  Neural Information Processing Systems 33: Annual Conference on Neural
  Information Processing Systems 2020, NeurIPS 2020, December 6-12, 2020,
  virtual}, 2020.

\bibitem{hafner2021mastering}
Danijar Hafner, Timothy Lillicrap, Mohammad Norouzi, and Jimmy Ba.
\newblock Mastering atari with discrete world models, 2021.

\bibitem{DBLP:conf/cvpr/GundavarapuSMSJ19}
Nitesh~B. Gundavarapu, Divyansh Srivastava, Rahul Mitra, Abhishek Sharma, and
  Arjun Jain.
\newblock Structured aleatoric uncertainty in human pose estimation.
\newblock In {\em {IEEE} Conference on Computer Vision and Pattern Recognition
  Workshops, {CVPR} Workshops 2019, Long Beach, CA, USA, June 16-20, 2019},
  pages 50--53. Computer Vision Foundation / {IEEE}, 2019.

\bibitem{girgis2021autobots}
Roger Girgis, Florian Golemo, Felipe Codevilla, Jim~Aldon D'Souza, Martin
  Weiss, Samira~Ebrahimi Kahou, Felix Heide, and Christopher Pal.
\newblock Autobots: Latent variable sequential set transformers, 2021.

\bibitem{DBLP:conf/cvpr/AlahiGRRLS16}
Alexandre Alahi, Kratarth Goel, Vignesh Ramanathan, Alexandre Robicquet,
  Fei{-}Fei Li, and Silvio Savarese.
\newblock Social {LSTM:} human trajectory prediction in crowded spaces.
\newblock In {\em 2016 {IEEE} Conference on Computer Vision and Pattern
  Recognition, {CVPR} 2016, Las Vegas, NV, USA, June 27-30, 2016}, pages
  961--971. {IEEE} Computer Society, 2016.

\bibitem{DBLP:conf/cvpr/HasanSTBGC18}
Irtiza Hasan, Francesco Setti, Theodore Tsesmelis, Alessio~Del Bue, Fabio
  Galasso, and Marco Cristani.
\newblock {MX-LSTM:} mixing tracklets and vislets to jointly forecast
  trajectories and head poses.
\newblock In {\em 2018 {IEEE} Conference on Computer Vision and Pattern
  Recognition, {CVPR} 2018, Salt Lake City, UT, USA, June 18-22, 2018}, pages
  6067--6076. {IEEE} Computer Society, 2018.

\bibitem{DBLP:conf/cvpr/Liang0NH019}
Junwei Liang, Lu~Jiang, Juan~Carlos Niebles, Alexander~G. Hauptmann, and
  Li~Fei{-}Fei.
\newblock Peeking into the future: Predicting future person activities and
  locations in videos.
\newblock In {\em {IEEE} Conference on Computer Vision and Pattern Recognition,
  {CVPR} 2019, Long Beach, CA, USA, June 16-20, 2019}, pages 5725--5734.
  Computer Vision Foundation / {IEEE}, 2019.

\bibitem{DBLP:conf/cvpr/LeeCVCTC17}
Namhoon Lee, Wongun Choi, Paul Vernaza, Christopher~B. Choy, Philip H.~S. Torr,
  and Manmohan Chandraker.
\newblock {DESIRE:} distant future prediction in dynamic scenes with
  interacting agents.
\newblock In {\em 2017 {IEEE} Conference on Computer Vision and Pattern
  Recognition, {CVPR} 2017, Honolulu, HI, USA, July 21-26, 2017}, pages
  2165--2174. {IEEE} Computer Society, 2017.

\bibitem{DBLP:conf/eccv/FelsenLG18}
Panna Felsen, Patrick Lucey, and Sujoy Ganguly.
\newblock Where will they go? predicting fine-grained adversarial multi-agent
  motion using conditional variational autoencoders.
\newblock In Vittorio Ferrari, Martial Hebert, Cristian Sminchisescu, and Yair
  Weiss, editors, {\em Computer Vision - {ECCV} 2018 - 15th European
  Conference, Munich, Germany, September 8-14, 2018, Proceedings, Part {XI}},
  volume 11215 of {\em Lecture Notes in Computer Science}, pages 761--776.
  Springer, 2018.

\bibitem{pmlr-v48-gal16}
Yarin Gal and Zoubin Ghahramani.
\newblock Dropout as a bayesian approximation: Representing model uncertainty
  in deep learning.
\newblock In Maria~Florina Balcan and Kilian~Q. Weinberger, editors, {\em
  Proceedings of The 33rd International Conference on Machine Learning},
  volume~48 of {\em Proceedings of Machine Learning Research}, pages
  1050--1059, New York, New York, USA, 20--22 Jun 2016. PMLR.

\bibitem{mangalam2020journey}
Karttikeya Mangalam, Harshayu Girase, Shreyas Agarwal, Kuan-Hui Lee, Ehsan
  Adeli, Jitendra Malik, and Adrien Gaidon.
\newblock It is not the journey but the destination: Endpoint conditioned
  trajectory prediction, 2020.

\bibitem{1618702}
T.~{Eltoft}, {Taesu Kim}, and {Te-Won Lee}.
\newblock On the multivariate laplace distribution.
\newblock {\em IEEE Signal Processing Letters}, 13(5):300--303, 2006.

\bibitem{Chang_2019_CVPR}
Ming-Fang Chang, John Lambert, Patsorn Sangkloy, Jagjeet Singh, Slawomir Bak,
  Andrew Hartnett, De~Wang, Peter Carr, Simon Lucey, Deva Ramanan, and James
  Hays.
\newblock Argoverse: 3d tracking and forecasting with rich maps.
\newblock In {\em Proceedings of the IEEE/CVF Conference on Computer Vision and
  Pattern Recognition (CVPR)}, June 2019.

\bibitem{DBLP:conf/icra/CuiRCLNHSD19}
Henggang Cui, Vladan Radosavljevic, Fang{-}Chieh Chou, Tsung{-}Han Lin, Thi
  Nguyen, Tzu{-}Kuo Huang, Jeff Schneider, and Nemanja Djuric.
\newblock Multimodal trajectory predictions for autonomous driving using deep
  convolutional networks.
\newblock In {\em International Conference on Robotics and Automation, {ICRA}
  2019, Montreal, QC, Canada, May 20-24, 2019}, pages 2090--2096. {IEEE}, 2019.

\bibitem{DBLP:journals/corr/abs-1808-05819}
Nemanja Djuric, Vladan Radosavljevic, Henggang Cui, Thi Nguyen, Fang{-}Chieh
  Chou, Tsung{-}Han Lin, and Jeff Schneider.
\newblock Motion prediction of traffic actors for autonomous driving using deep
  convolutional networks.
\newblock {\em CoRR}, abs/1808.05819, 2018.

\bibitem{DBLP:journals/corr/abs-2008-08294}
Hang Zhao, Jiyang Gao, Tian Lan, Chen Sun, Benjamin Sapp, Balakrishnan
  Varadarajan, Yue Shen, Yi~Shen, Yuning Chai, Cordelia Schmid, Congcong Li,
  and Dragomir Anguelov.
\newblock {TNT:} target-driven trajectory prediction.
\newblock {\em CoRR}, abs/2008.08294, 2020.

\bibitem{DBLP:conf/icra/MercatGZSBG20}
Jean Mercat, Thomas Gilles, Nicole~El Zoghby, Guillaume Sandou, Dominique
  Beauvois, and Guillermo~Pita Gil.
\newblock Multi-head attention for multi-modal joint vehicle motion
  forecasting.
\newblock In {\em 2020 {IEEE} International Conference on Robotics and
  Automation, {ICRA} 2020, Paris, France, May 31 - August 31, 2020}, pages
  9638--9644. {IEEE}, 2020.

\bibitem{nuscenes2019}
Holger Caesar, Varun Bankiti, Alex~H. Lang, Sourabh Vora, Venice~Erin Liong,
  Qiang Xu, Anush Krishnan, Yu~Pan, Giancarlo Baldan, and Oscar Beijbom.
\newblock nuscenes: A multimodal dataset for autonomous driving.
\newblock {\em arXiv preprint arXiv:1903.11027}, 2019.

\bibitem{kingma2014method}
Diederik~P. Kingma and Jimmy Ba.
\newblock Adam: A method for stochastic optimization, 2014.
\newblock cite arxiv:1412.6980Comment: Published as a conference paper at the
  3rd International Conference for Learning Representations, San Diego, 2015.

\end{thebibliography}
}
\clearpage

\section*{Appendix}
\appendix
\section{Proof of Laplace Model Design}
\begin{proof}
Consider the $i$-th data sample $(\X^i,\Y^i)$, as in the training process of the prediction model the values of $\X^i$ and $\Y^i$ are given, $p(\Y^i|\z^i,\X^i;\w)$ is a function of $\z^i \in \R^{+}$ with the probability density function: $p(\z^i|\X^i) = \frac{1}{\lambda}e^{-\frac{\z^i}{\lambda}}$:
\begin{align*}
p(\Y^i|\z^i,\X^i;\w)&=f_{\w}(\z^{i}) = \frac{1}{(\z^{i})^{\frac{m}{2}}}e^{-\frac{g_{\w}^{i}}{\z^{i}}},
\end{align*}
where $g_{\w}^{i} = \frac{1}{2}(\Y^{i} - \mu_{\w}(\X^{i}))[\Sigma_{\w}^{-1}(\X^{i})](\Y^{i} - \mu_{\w}(\X^{i}))^{T}$ and $m\in\NN^{+}$ is the number of the agents in the $i$-th data sample. We need to prove that there should exist a $(\z^{i})^{*} \in\R^{+}$ to make:
\begin{align*}
p(\Y^i|\X^i;\w) &=\int_{0}^{+\infty} p(\Y^i|\z^i,\X^i;\w)p(\z^i|\X^i;\w) d\z^i\nonumber\\
&=\int_{0}^{+\infty} f_{\w}(\z^i)p(\z^i|\X^i) d\z^i\nonumber\\
&=E_{\z^i}[f_{\w}(\z^i)]\nonumber\\
&=f_{\w}((\z^{i})^{*})\nonumber\\
&=p(\Y^{i}|(\z^{i})^{*},\X^{i};\w).
\end{align*}
And the existence of $(\z^i)^*$ can be proved by proving a fact that, when $\z^{i}\in\R^{+}$, $f(\z^{i})$ is a continuous bounded function.

As the $g^{i}_{\w}$ can be reformulated as:
\begin{align}
g_{\w}^{i} &= \frac{1}{2}(\Y^{i} - \mu_{\w}(\X^{i}))[L^{\prime}_{\w}(\X)L'^{T}_{\w}(\X)](\Y^{i} - \mu_{\w}(\X^{i}))^{T}\nonumber\\
&=\frac{1}{2}[(\Y^{i} - \mu_{\w}(\X^{i}))L^{\prime}_{\w}(\X)][(\Y^{i} - \mu_{\w}(\X^{i}))L^{\prime}_{\w}(\X)]^{T}\nonumber\\
&\geq 0,
\label{nq}
\end{align}
where $L^{\prime}_{\w}(\X)$ is a lower triangular matrix and the equal sign of (\ref{nq}) is only true when $\mu_{\w}(\X^{i})$ is equal to $\Y^{i}$, but in practice, $\mu_{\w}(\X^{i})$ is hardly equal to $\Y^{i}$, which means in the training process we have:
\begin{align}
g_{\w}^{i}~\textgreater~0.
\label{g_greater}
\end{align}

Based on (\ref{g_greater}), let $\s^i = \frac{1}{\z^i}$, then as $\z^{i}\to0^{+}$, we have $\s^{i}\to+\infty$, so for $\z^{i}\to0^{+}$:
\begin{align}
\lim\limits_{\z^{i}\to0^{+}}f_{\w}(\z^{i}) &=  \lim\limits_{\s^{i}\to+\infty}(\s^i)^{\frac{m}{2}} e^{-\s^i g_{\w}^{i}}\nonumber\\
&= \lim\limits_{\s^{i}\to+\infty}\frac{(\frac{m}{2})!}{(g_{\w}^{i})^{\frac{m}{2}}e^{\s^i g_{\w}^{i}}}\nonumber\\
&=0
\label{lim2}
\end{align}

For $\z^{i}\to +\infty$:
\begin{equation}
\lim\limits_{\z^{i}\to+\infty}f_{\w}(\z^{i}) = 0
\label{lim}
\end{equation}

Furthermore, as the derivative of $f_{\w}(\z^{i})$ is then:
\begin{equation}
\setlength{\abovedisplayskip}{4pt}
\setlength{\belowdisplayskip}{4pt}
f^{\prime}_{\w}(\z^{i}) = (\z^i)^{-\frac{m}{2}-2}\cdot e^{-(\z^i)^{-1} g_{\w}^{i}}\cdot(g_{\w}^{i}-\frac{m}{2}\z^{i}).
\label{de}
\end{equation}
According to (\ref{g_greater}), (\ref{lim}), (\ref{lim2}) and (\ref{de}), when we set $f^{\prime}_{\w}(\z^{i}) = 0$, we can get the maximum value of $f_{\w}(\z^{i})$ is  $f_{\w}(\frac{2g_{\w}^{i}}{m})\in\R^+$.

On the basis of above discussions, when $\z^{i}\in\R^{+}$, $f(\z^{i})$ is a continuous bounded function, which means the $(\z^{i})^{*}$ is existent.
\end{proof}

\section{Toy Problem}

\subsection{Generation Details of Synthetic Datasets}
\begin{figure}[h]
\begin{center}
\centerline{\subfigure[$\mu_{gt}$]{
         \label{fig:independencea}
 		\includegraphics[scale=0.33]{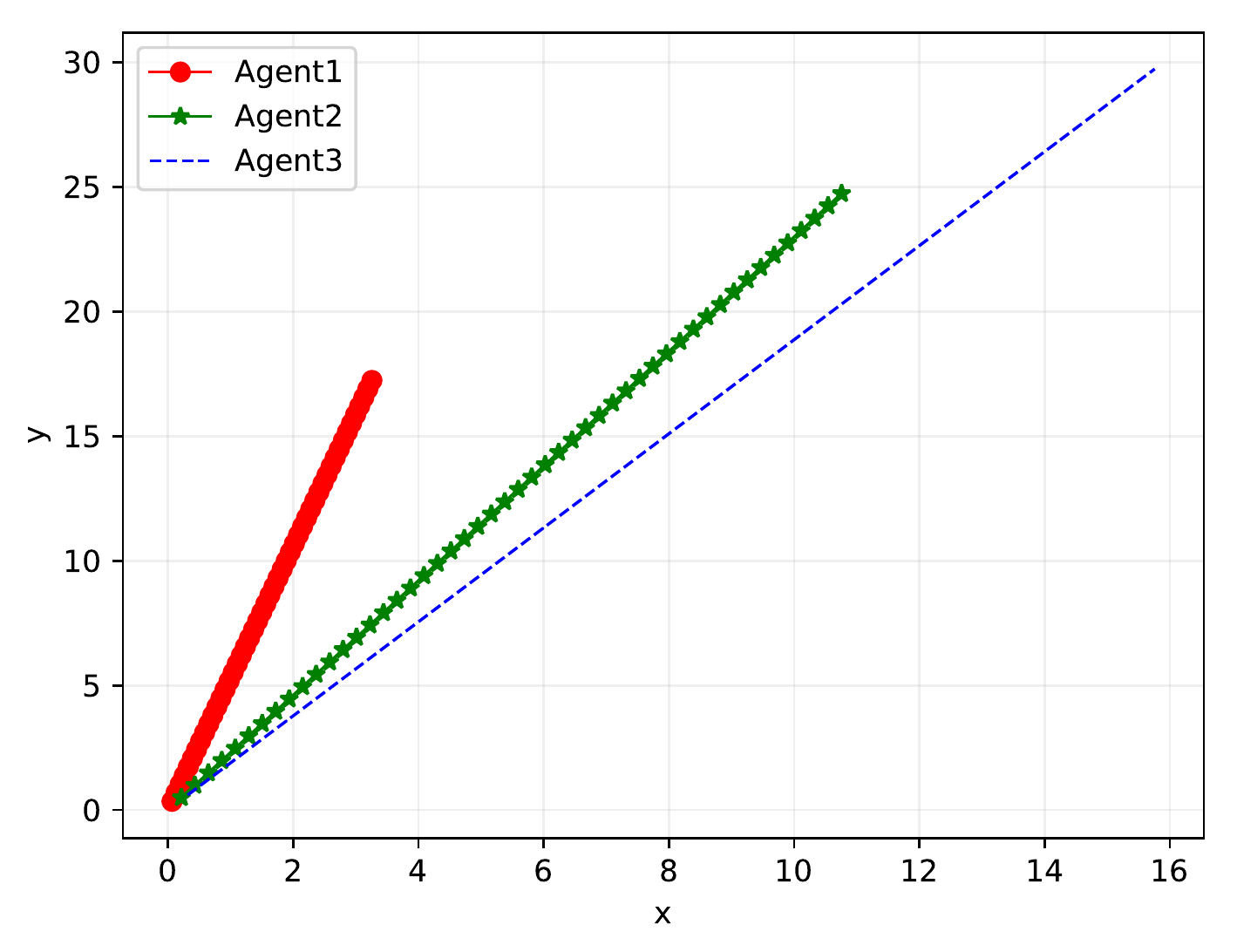}
 	}
 \subfigure[$\epsilon$]{
         \label{fig:independenceb}
 		\includegraphics[scale=0.33]{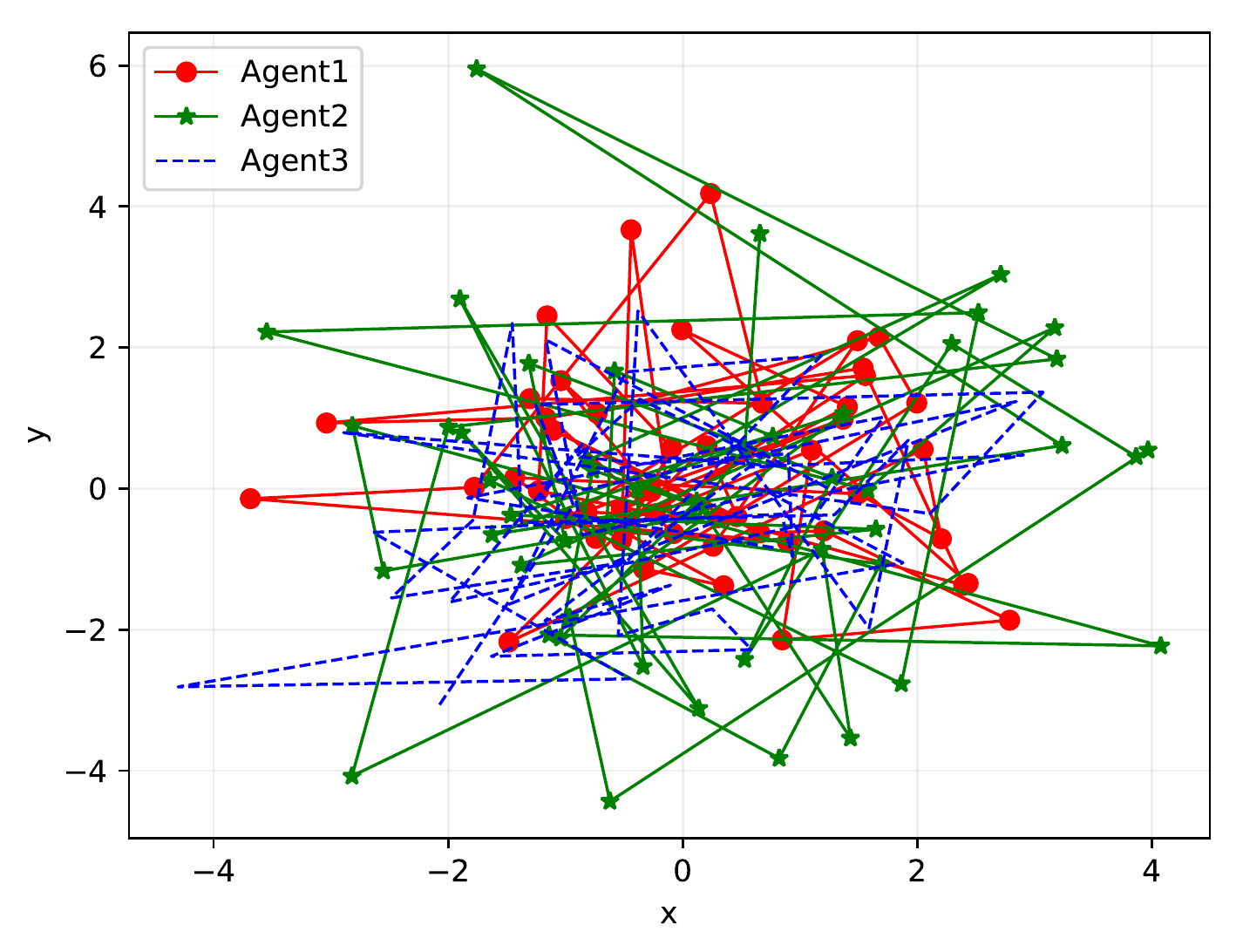}
 	}
 \subfigure[$\x$]{
         \label{fig:independenceb}
		\includegraphics[scale=0.33]{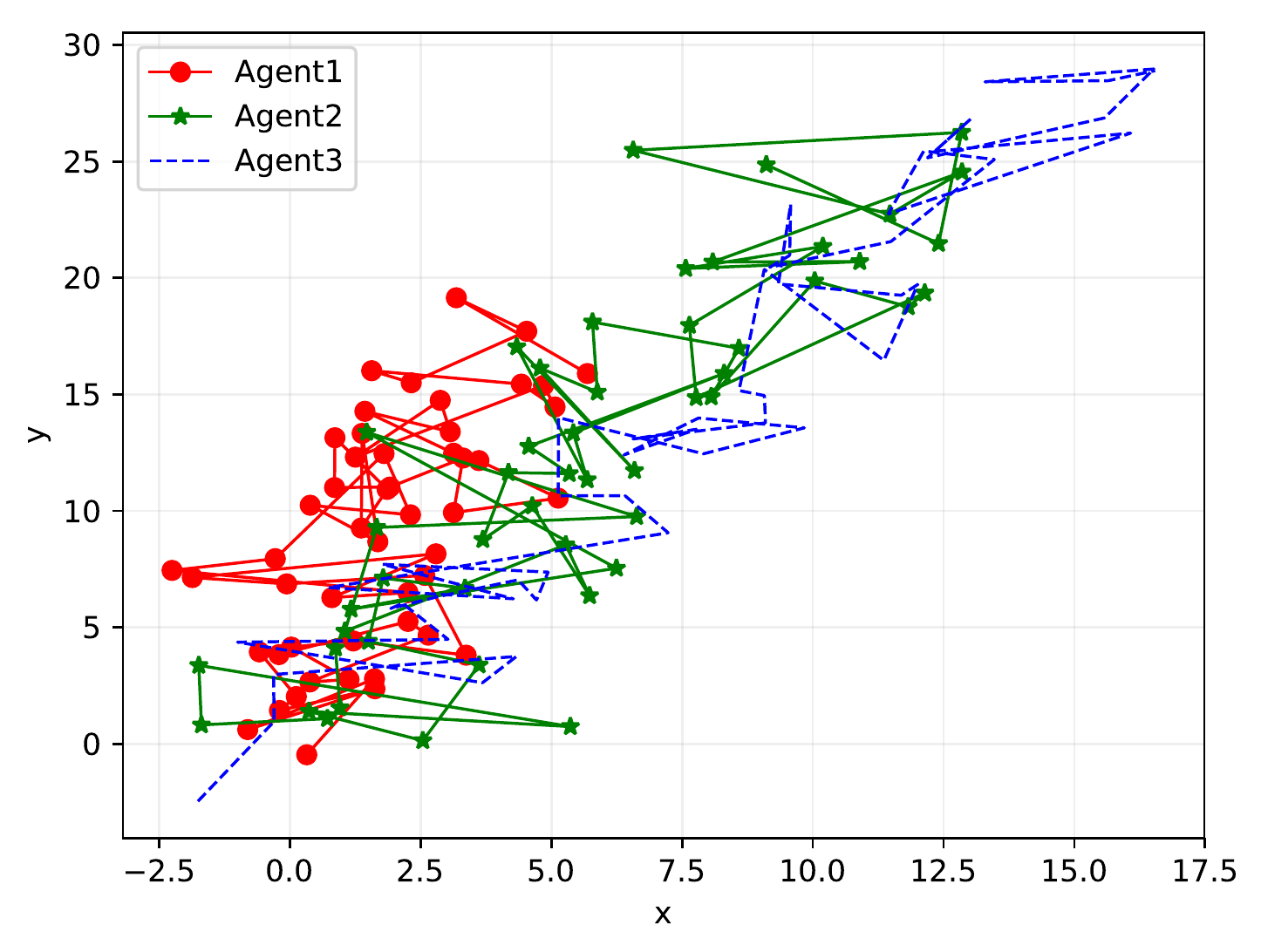}
	}	
	}
\caption{\textbf{Sample visualization of the generation of Gaussian synthetic dataset.} $\x$ is the sum of $\mu_{gt}$ and $\epsilon$.}
\label{quali_sim2}
\vspace{-2em}
\end{center}
\end{figure}

\begin{figure}[h]
\begin{center}
\centerline{\subfigure[$\mu_{gt}$]{
         \label{fig:independencea}
 		\includegraphics[scale=0.33]{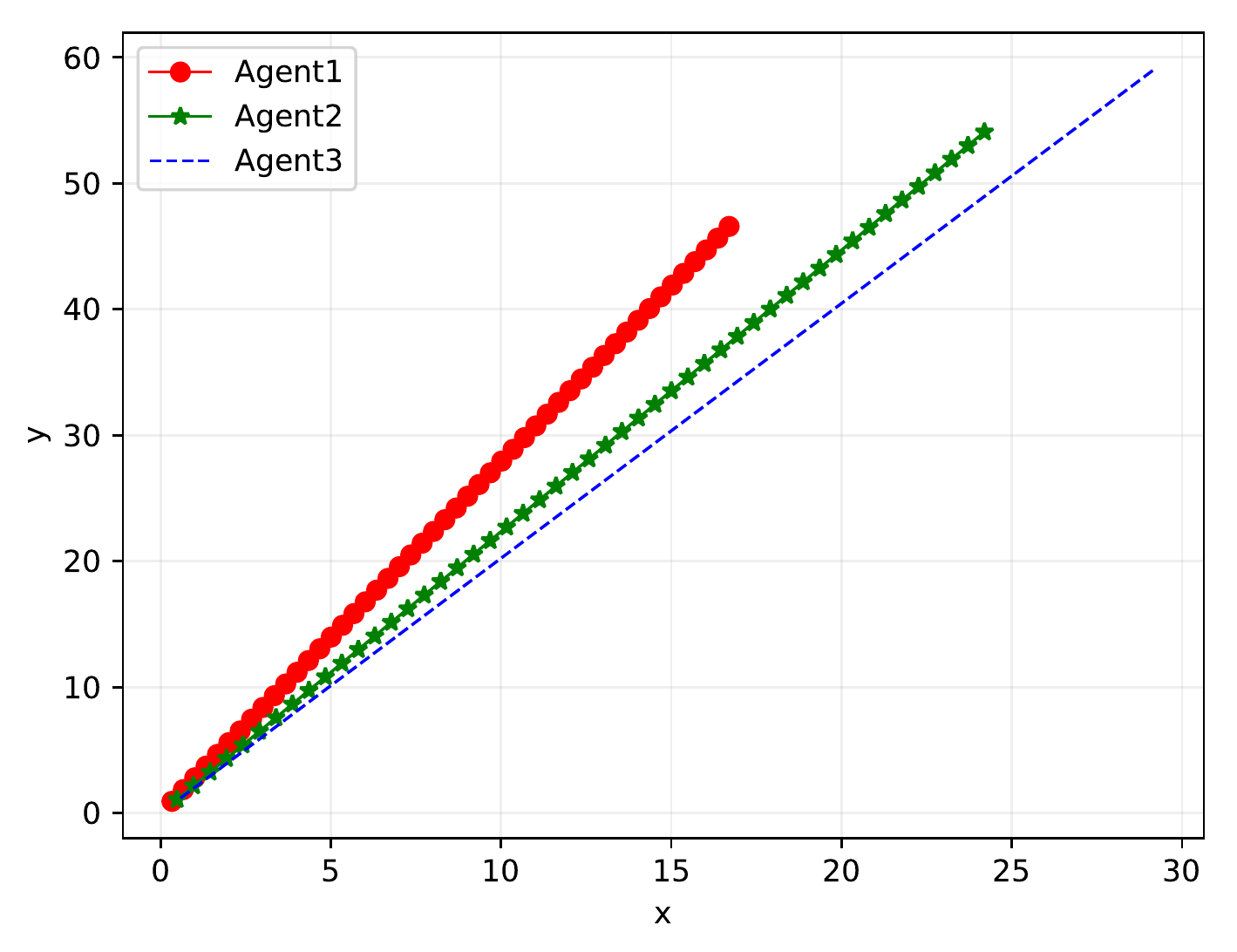}
 	}
 \subfigure[$\epsilon$]{
         \label{fig:independenceb}
 		\includegraphics[scale=0.33]{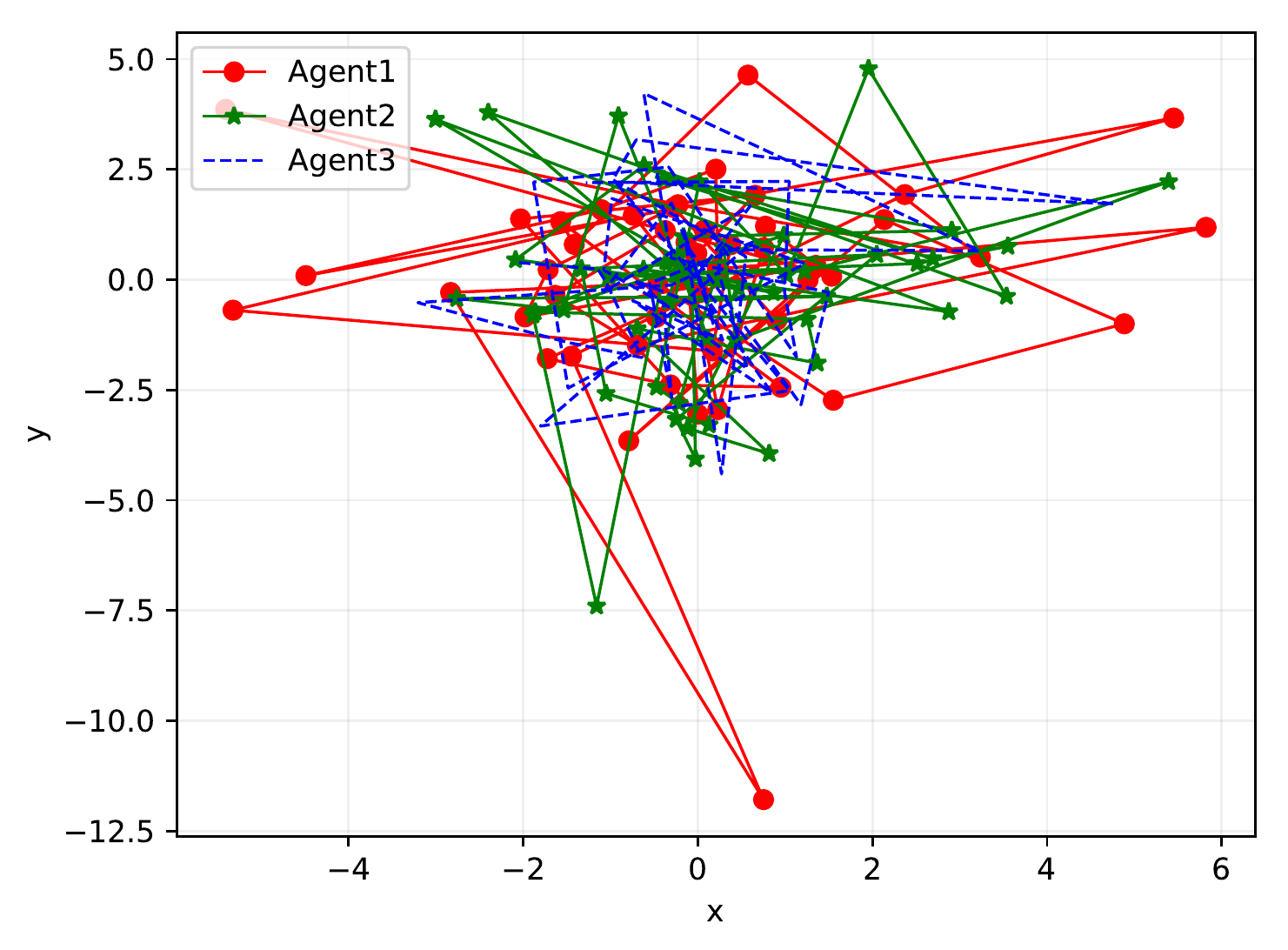}
 	}
 \subfigure[$\x$]{
         \label{fig:independenceb}
		\includegraphics[scale=0.33]{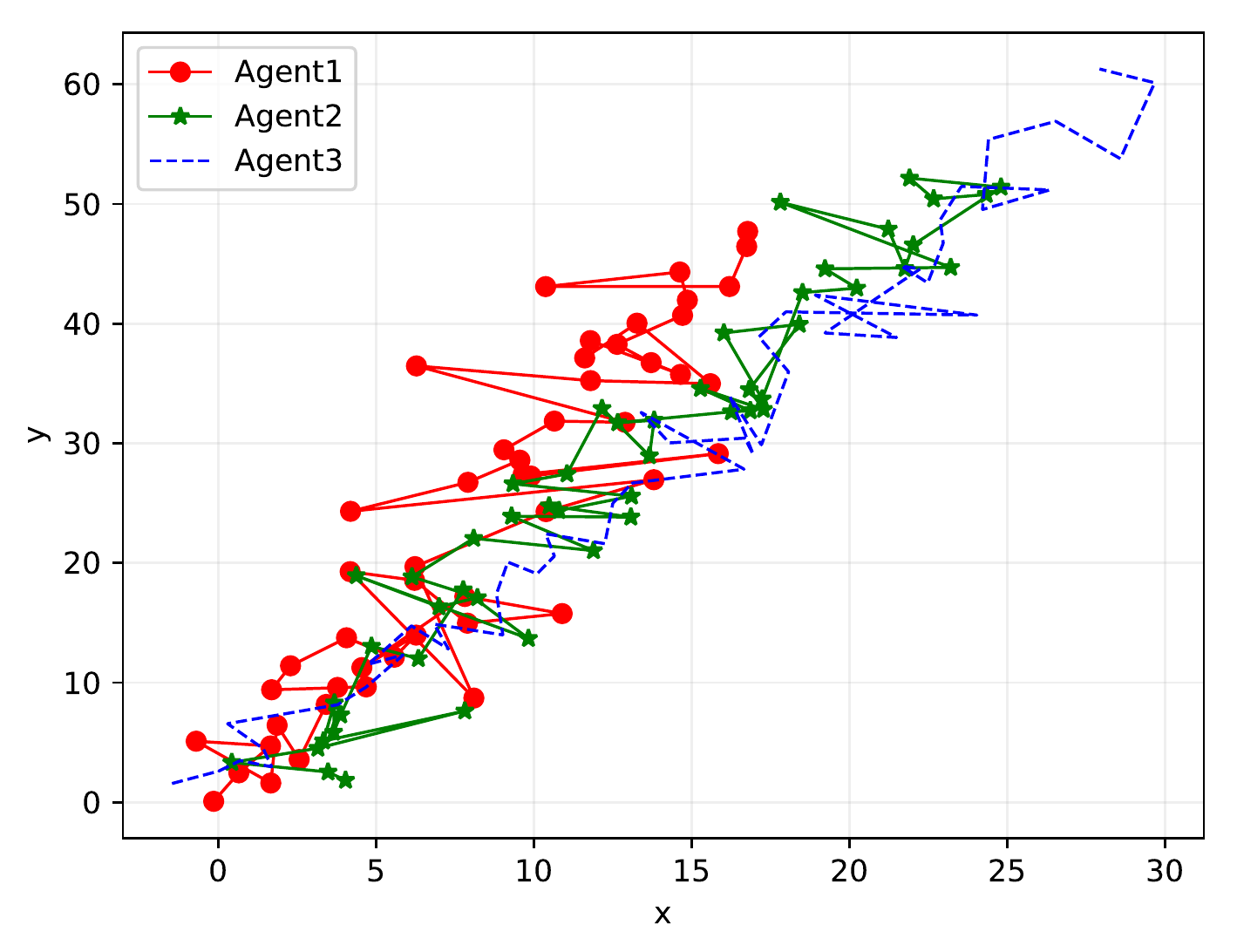}
	}	
	}
\caption{\textbf{Sample visualization of the generation of Laplace synthetic dataset.} $\x$ is the sum of $\mu_{gt}$ and $\epsilon$.}
\vspace{-2em}
\end{center}
\end{figure}

For the Gaussian synthetic dataset, since a random variable $\x$ that obeys a multivariate Gaussian distribution $\N(\mu,\Sigma)$ can be formulated as the sum of the mean $\mu$ and a random variable $\epsilon$: $x = \mu + \epsilon$, where $\epsilon \sim \N(0,\Sigma)$. For generating the Gaussian synthetic dataset, we firstly generate 50 different two-dimensional coordinates of three agents that move in a uniform straight line. We denote this part of the data as $\mu_{gt}$, and set it as the mean of the multivariate Gaussian distribution to which the trajectories belong. Subsequently, we sample a set of data $\epsilon$ from a multivariate Gaussian distribution $\N(0,\Sigma_{gt})$ (where $\Sigma_{gt} \in \R^{3\times3}$), obviously this set of data contains the information of the covariance matrix of its distribution. Finally, we add the data $\mu_{gt}$ representing the mean value of the distribution and the data $\epsilon$ representing the covariance matrix information of the distribution to get our final data $\x$, which is $\x = \mu_{gt} + \epsilon$. At this time, the data $\x$ we get is equivalent to the data sampled from the multivariate Gaussian distribution $\N(\mu_{gt},\Sigma_{gt})$. Moreover, following similar steps, we can get the Laplace synthetic dataset.

\subsection{Metric Computation Details}
{$\ell_{2}$ of $\mu$} is the average of pointwise $\ell_2$ distances between the estimated mean and the ground truth mean. {$\ell_1$ of $\Sigma$} is the average of pointwise $\ell_1$ distances between the estimated covariance matrix and the ground truth covariance.

KL is the KL divergence between the ground truth distribution and the estimated distribution $D_{KL}(p_{g}(\!X\!)||p_{e}(\!X\!))$, where $p_{e}(\!X\!)\sim\N(\mu_{p_{e}},\Sigma_{p_{e}})$ is the estimated distribution, $p_{g}(\!X\!)\sim\N(\mu_{p_{g}},\Sigma_{p_{g}})$ is the ground truth distribution, $\Sigma_{p_{e}} \in \R^{k\times k}$ and $\Sigma_{p_{g}} \in \R^{k\times k}$ . For multivariate Gaussian distribution, we compute it by the following formula~(\ref{kl_gau}):
\begin{align}
D_{KL}(p_{g}(\!X\!)||p_{e}(\!X\!)) &=p_{g}(\!X\!)\int_{X}[\log(p_{g}(\!X\!)) - \log(p_{e}(\!X\!))]dX\nonumber\\
&= \frac{1}{2}[\log(\frac{|\Sigma_{p_{e}}|}{|\Sigma_{p_{g}}|})-k+(\mu_{p_{g}} - \mu_{p_{e}})^{T}\Sigma_{p_{e}}^{-1}(\mu_{p_{g}} - \mu_{p_{e}}) + trace\{\Sigma_{p_{e}}^{-1}\Sigma_{p_{g}}\}].
\label{kl_gau}
\end{align}
For multivariate Laplace distribution, as the probability density function of it is too complicated, when we compute the KL divergence, we firstly compute the value of $p_{g}(\!X\!)$ and $p_{e}(\!X\!)$ for each given data sample $X$ respectively, and then we compute $D_{KL}(p_{g}(\!X\!)||p_{e}(\!X\!))$ by the following formula~(\ref{kl_lap}):
\begin{align}
D_{KL}(p_{g}(\!X\!)||p_{e}(\!X\!)) &=\sum_{X}p_{g}(\!X\!)[\log(p_{g}(\!X\!)) - \log(p_{e}(\!X\!))].
\label{kl_lap}
\end{align}

\subsection{Implementation Details}
\begin{figure}[h]
\begin{center}
\vskip -0.12in
\subfigure[Framework]{
        \label{sim_ex_fr}
		\includegraphics[width=0.5\columnwidth]{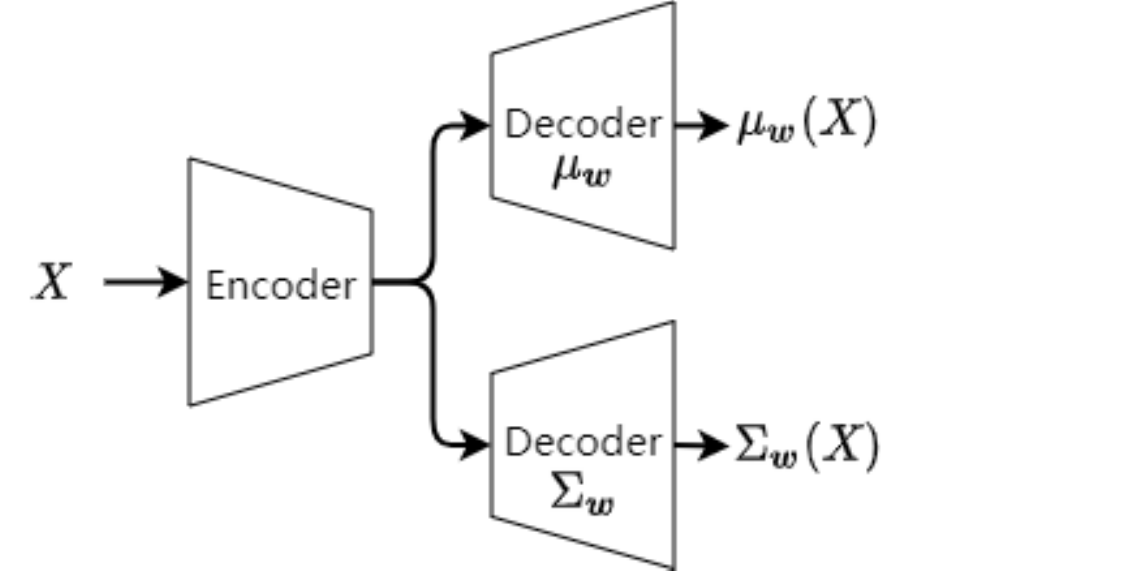}
	}
\subfigure[Four-Layer MLP]{
        \label{sim_ex_mlp}
		\includegraphics[width=0.8\columnwidth]{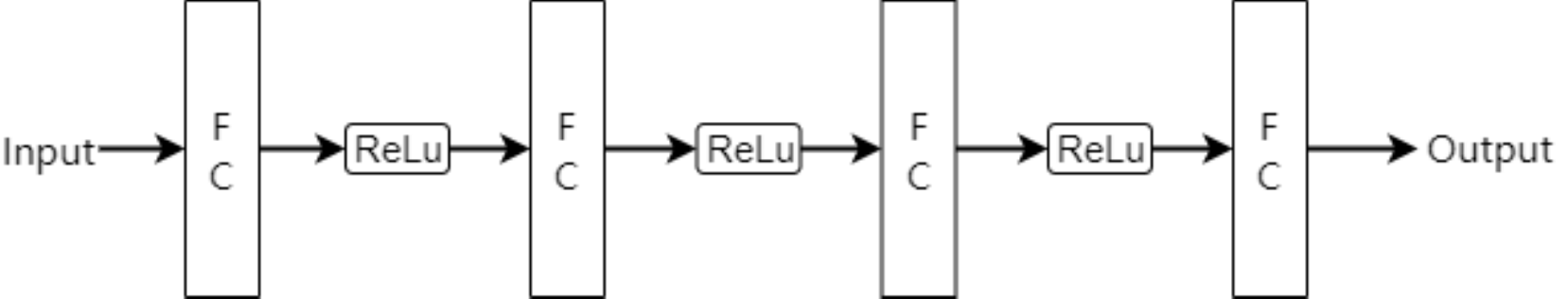}
	}
\caption{\textbf{The network architecture used in synthetic datasets. }(a): The framework of the used network. (b): The four-layer multilayer perceptron used to form the encoder and decoders of the used network, where FC denotes the full connected layer and ReLu denotes the ReLu activation function.}
\label{sim_use}
\end{center}
\end{figure}

\textbf{Model Structure:} For toy problem, the network architecture contains an encoder and two decoders, all of which are four-layer multilayer perceptrons (MLPs), which are shown in Figure~\ref{sim_use}.

\textbf{Training Details:} For toy problem, we train the model on 1 GTX 1080Ti GPU using a batch size of 72 with the Adam~\cite{kingma2014method} optimizer with an initial learning rate of $5\times10^{-3}$ and the training process finishes at 36 epochs.



\section{Real World Problem}

\subsection{Implementation Details}
\begin{figure}[tbh!]
    \centering
    \includegraphics[scale=0.6]{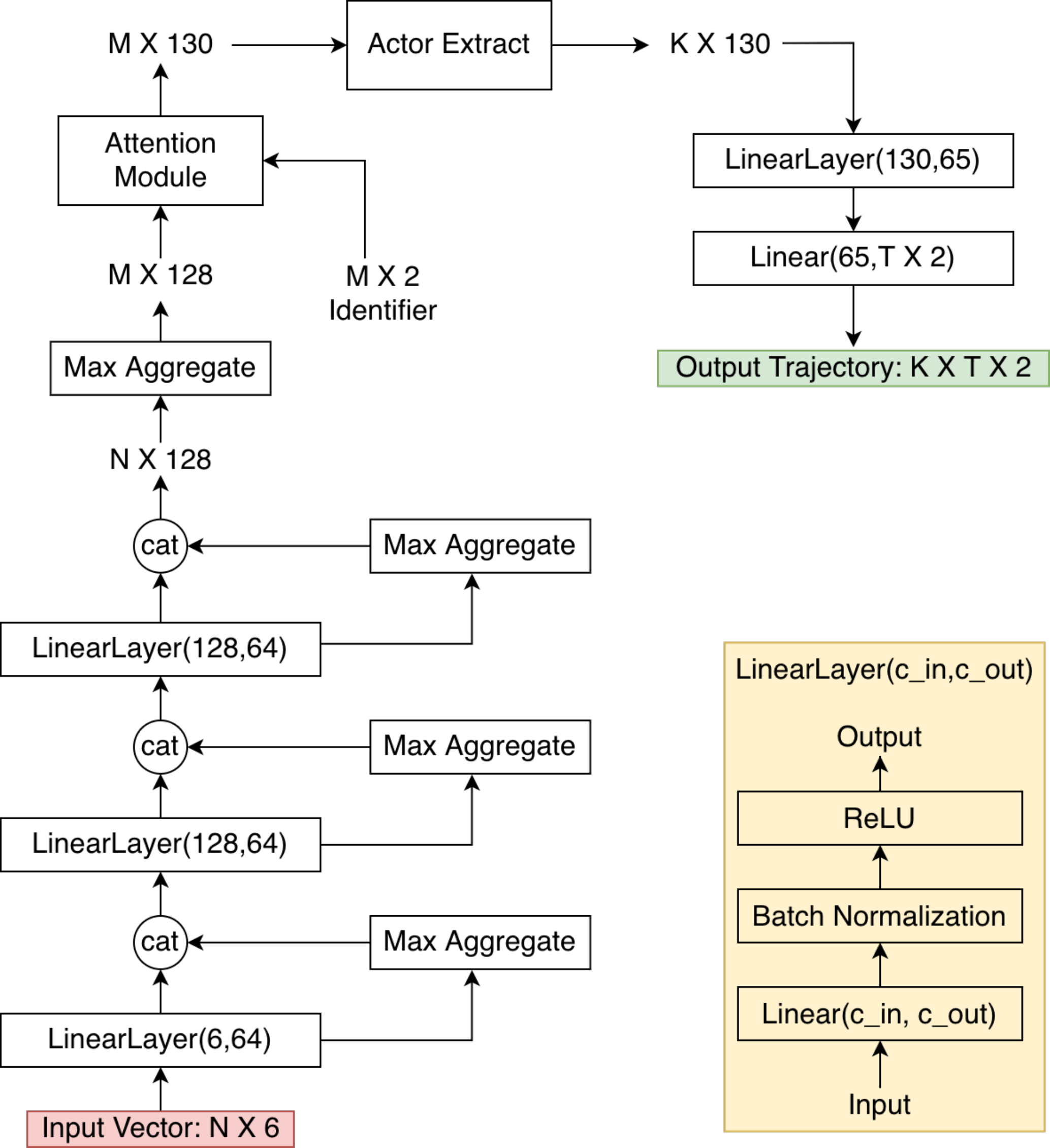}
    \caption{\textbf{Implementation structure of VectorNet.} N is the input vector number, M is the polygon number, K is the actor number.More details please refer to \cite{Gao_2020_CVPR}}
    \label{fig:vectornet}
\end{figure}
\textbf{Model Structure:} We implemented the LaneGCN according to the structure shown in the appendix of \cite{liang2020learning}. And the structure of our implemented VectorNet is shown in Figure \ref{fig:vectornet}.

\textbf{Training Details:} For LaneGCN in Argoverse, we train the model on 1 GTX 1080Ti GPU using a batch size of 64 with the Adam~\cite{kingma2014method} optimizer with an initial learning rate of $2.5\times10^{-4}$ and the training process finishes at 65 epochs. For LaneGCN in nuScenes, we train the model on 2 GTX 1080Ti GPUs using a batch size of 56 with the Adam~\cite{kingma2014method} optimizer with an initial learning rate of $8\times10^{-4}$ and the training process finishes at 90 epochs. For VectorNet in Argoverse, we train the model on 2 GTX 1080Ti GPUs using a batch size of 128 with the Adam~\cite{kingma2014method} optimizer with an initial learning rate of $2\times10^{-3}$ and the training process finishes at 105 epochs. For VectorNet in nuScenes, we train the model on 2 GTX 1080Ti GPUs using a batch size of 128 with the Adam~\cite{kingma2014method} optimizer with an initial learning rate of $1\times10^{-3}$ and the training process finishes at 500 epochs.

\begin{table*}[!h]
\caption{Two special cases with various assumptions about covariance $\Sigma$. \textbf{ID} denotes the identity matrix (no uncertainty). \textbf{DIA} denotes the diagonal matrix (individual uncertainty). \textbf{FULL} denotes the full matrix (individual and collaborative uncertainty).}
\vspace{-1em}
\begin{center}
\begin{scriptsize}
\begin{sc}
\begin{tabular}{c|c|c}
\toprule
\multirow{2}{*}{Assumption} &  \multicolumn{2}{c}{Two Special Cases}
\\
\cline{2-3}
&Gaussian Distribution & Laplace Distribution \\
\hline
\makecell[c]{ID$:
\begin{pmatrix}
 \!1\!&\!0  \!&\!\cdots\!&\!0\!\\
 \!0\!&\!1\!&\!\cdots\!&\!0\!\\
 \!\vdots\!&\!\vdots\!&\!&\!\vdots\!\\
 \!0\!&\!0\!&\!\cdots\!&1\!\\
\end{pmatrix}$}\!&\!\makecell[c]{$||\Y\!-\!\mu_{\w}(\X)||_2^{2}$}& \makecell[c]{$||\Y\!-\!\mu_{\w}(\X)||_1$}
\\
\hline
DIA$:
\begin{pmatrix}
\sigma_{11} \!& \!0\!&\!\cdots\!&\!0\\
0\!&\!\sigma_{22}\!&\!\cdots\!&\!0\\
\vdots\!&\!\vdots\!&\!&\!\vdots  \\
0\!&\!0\!&\!\cdots\!&\!\sigma_{mm}  \\
\end{pmatrix}$\!

&{\scriptsize\!$\!\frac{1}{2}\! \sum\limits_{i=1}^{m}\![\sigma_{ii}^{-2}||\!\y_i\!-\!\mu_{\w}(\!\x_i\!)\!||_2^{2}\!+\!\log\!\sigma_{ii}^{2}]$\!}

&{\scriptsize\!$\!\sum\limits_{i=1}^{m}[\sigma_{ii}^{-2}||\!\y_i\!-\!\mu_{\w}(\!\x_i\!)\!||_1\!+\!\log\!\sigma_{ii}^{2}]$}
\\
\hline
\makecell[c]{FULL$:
\begin{pmatrix}
\sigma_{11}\!&\!\sigma_{12}\!&\!\cdots\!& \sigma_{1m}\\
\sigma_{21}\!&\!\sigma_{22}\!&\!\cdots\!&\!\sigma_{2m}  \\
\vdots\!&\!\vdots\!&\!&\!\vdots\\
\sigma_{m1}\!&\!\sigma_{m2}\!&\!\cdots\!&\!\sigma_{mm}\\
\end{pmatrix}$}\!

&{\scriptsize\!\makecell[c]{$\!\frac{1}{2} [q_{\w}(\Y,\X)\!-\sum\limits_{j=1}^{m}\!\log (d_{jj})]$} }

& {\scriptsize \makecell[c]{$ \frac{1}{2} [\frac{q_{\w}(\Y,\X)}{\Phi_{\w}(\X)}+m\log \Phi_{\w}(\X)-
    \!\sum\limits_{j=1}^{m}\!\log (d_{jj})]
    $}}
\\
\bottomrule
\end{tabular}
\end{sc}
\end{scriptsize}
\end{center}
\vspace{-0.2in}
\end{table*}

\begin{table*}[tbh]
\caption{Ablation on assumptions about covariance $\Sigma$ of chosen probability density functions (PDFs) in single future prediction. \textbf{ID} denotes the identity matrix (no uncertainty). \textbf{DIA} denotes the diagonal matrix (individual uncertainty). \textbf{FULL} denotes the full matrix  (individual and collaborative uncertainty). On Argoverse and nuScenes, a model with individual uncertainty surpasses a model without uncertainty; a model with individual and collaborative uncertainty surpasses a model with individual uncertainty only.}
\label{compa_id}
\vspace{-2mm}
\vskip -0.5in
\begin{center}
\begin{footnotesize}
\begin{small}
\begin{sc}
\begin{tabular}{c|c|c|cc|cc}
\toprule
\multirow{3}{*}{Dataset} & \multirow{3}{*}{Method} & \multirow{3}{*}{Assumption about $\Sigma$} & \multicolumn{4}{c}{Type of Chosen PDF}
\\
&&  & \multicolumn{2}{c|}{Gaussian}&\multicolumn{2}{c}{Laplace}
\\
& &  & ADE& FDE& ADE& FDE
\\
\hline
\multirow{6}{*}{\scriptsize Argoverse} & \multirow{3}{*}{\scriptsize LaneGCN} & ID & 1.52 & 3.32 & 1.44 & 3.17
\\
&& DIA & 1.45 & 3.19 & 1.43 & 3.16
\\
&&FULL & \textbf{1.42} & \textbf{3.14} & \textbf{1.41} & \textbf{3.11}
\\
\cline{2-7}
& \multirow{3}{*}{\scriptsize VectorNet} & ID & 1.67 & 3.62 & 1.59 & 3.44
\\
&& DIA &1.63 & 3.60 & 1.56 & 3.42
\\
&& FULL &  \textbf{1.57} & \textbf{3.46} & \textbf{1.52} & \textbf{3.34}
\\
\hline
\hline
\multirow{6}{*}{\scriptsize nuScenes}&\multirow{3}{*}{\scriptsize LaneGCN} & ID&  4.50 & 10.62 & 4.47 & 10.54
\\
&&DIA&  4.47 & 10.59 & 4.34 & 10.34
\\
&&FULL& \textbf{4.39} & \textbf{10.44} & \textbf{4.25} & \textbf{10.15}
\\
\cline{2-7}
&\multirow{3}{*}{\scriptsize VectorNet} & ID &4.23 & 9.91 & 4.09 & 9.80
\\
&&DIA & 4.07 & 9.86 & 4.02 & 9.79
\\
&& FULL & \textbf{3.99} & \textbf{9.57} & \textbf{3.81} & \textbf{9.22}
\\
\bottomrule
\end{tabular}
\end{sc}
\end{small}
\end{footnotesize}
\end{center}
\vskip -0.1in
\end{table*}

\begin{table*}[tbh]
\caption{Ablation on assumptions about covariance $\Sigma$ of chosen probability density functions (PDFs) on the basis of the official version of LaneGCN in single future prediction. \textbf{ID} denotes the identity matrix (no uncertainty). \textbf{DIA} denotes the diagonal matrix (individual uncertainty). \textbf{FULL} denotes the full matrix  (individual and collaborative uncertainty). On both of the validate set and the test set of Argoverse, a model with individual uncertainty surpasses a model without uncertainty; a model with individual and collaborative uncertainty surpasses a model with individual uncertainty only.}
\label{compa_id5}
\vspace{-2mm}
\vskip -0.5in
\begin{center}
\begin{footnotesize}
\begin{small}
\begin{sc}
\begin{tabular}{c|c|c|cc|cc}
\toprule
\multirow{3}{*}{Dataset} & \multirow{3}{*}{Set} & \multirow{3}{*}{Assumption about $\Sigma$} & \multicolumn{4}{c}{Type of Chosen PDF}
\\
&&  & \multicolumn{2}{c|}{Gaussian}&\multicolumn{2}{c}{Laplace}
\\
& &  & ADE& FDE& ADE& FDE
\\
\hline
\multirow{6}{*}{\scriptsize Argoverse} & \multirow{3}{*}{\scriptsize Validate} & ID & 1.36 & 2.94 & 1.28 & 2.78
\\
&& DIA & 1.32 & 2.90 & 1.27 & 2.76
\\
&&FULL & \textbf{1.31} & \textbf{2.89} & \textbf{1.26} & \textbf{2.75}
\\
\cline{2-7}
& \multirow{3}{*}{\scriptsize Test} & ID & 1.69 & 3.72 & 1.64 & 3.61
\\
&& DIA &1.67 & 3.70 & 1.62 & 3.56
\\
&& FULL &  \textbf{1.66} & \textbf{3.67} & \textbf{1.61} & \textbf{3.53}
\\
\bottomrule
\end{tabular}
\end{sc}
\end{small}
\end{footnotesize}
\end{center}
\vskip -0.1in
\end{table*}

\subsection{Additional Results}
We compare our proposed approach with the approach not modeling uncertainty, which assumes the covariance $\Sigma$ is an identity matrix (ID). As the results illustrated in Table~\ref{compa_id}, our proposed Laplace CU-based framework still enables LaneGCN and VectorNet to achieve the best performances on ADE \& FDE metrics on both Argoverse and nuScenes benchmarks in single future prediction.

After the paper is accepted in NeurIPS 2021, we also apply our proposed framework to the official version of LaneGCN (with map information) and test it on the Argoverse benchmark in single future prediction. As results shown in the Table~\ref{compa_id5}, our proposed Laplace CU-based framework still enables LaneGCN to achieve the best performances on ADE \& FDE metrics on both validate set and test set of Argoverse benchmarks in single future prediction.

\subsection{Extra Visualization Results}
\begin{figure}[tbh!]
\begin{center}
\includegraphics[width=1.\columnwidth]{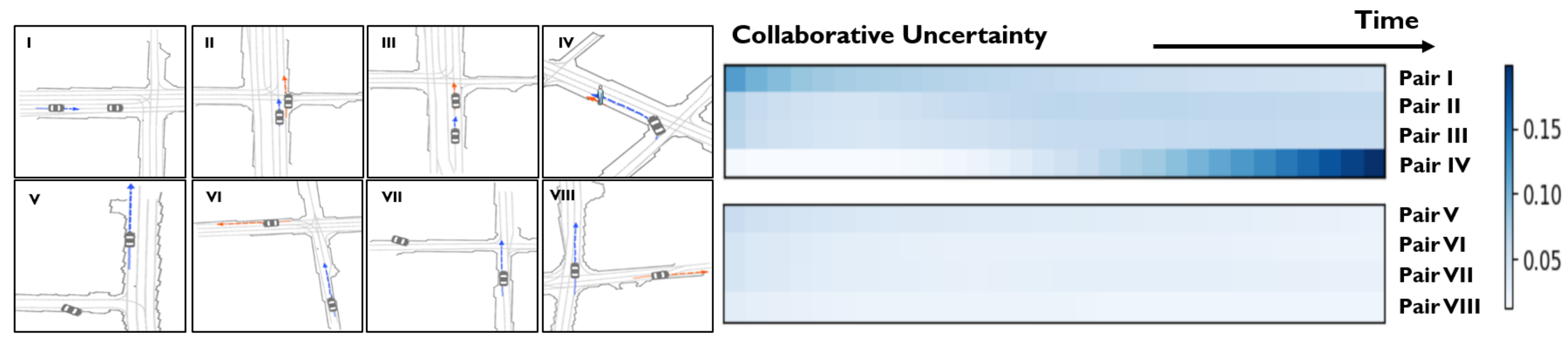}

\caption{\textbf{Visualization of CU on Argoverse dataset}. Pair I: One agent approaching another agent parking at an intersection waiting for green light, as little new interactive information would be generated before the red light turns green, CU decreases over time. Pair II: Agents moving side by side, which might generate complicated interactive information making CU show a non-monotonic change over time. Pair III: Agents driving on the same road, which might generate complicated interactive information making CU show a non-monotonic change over time. Pair IV: Agents moving close to each other, CU increases over time. Pair V to Pair VIII: Agents located in completely different areas on the map, CUs are close to zero.}
\label{Visualizationap}
\end{center}
\end{figure}

\textbf{Visualization of collaborative uncertainty between two agents.} In Figure~\ref{Visualizationap}, there are 8 actor pairs (blue and orange lines) trajectories (solid lines are the past trajectories and dashed lines are the future trajectories) and their corresponding collaborative uncertainty values changing over the last 30 frames (the heatmap). Pair I to IV and Pair V to VIII are the ones w/o obvious interaction. These results show that the value of collaborative uncertainty is highly related to the amount of the interactive information among agents.

\end{document}